\PassOptionsToPackage{table}{xcolor}
\documentclass[acmsmall,screen]{acmart}

\AtBeginDocument{%
  }

\setcopyright{acmcopyright}
\copyrightyear{2018}
\acmYear{2018}
\acmDOI{XXXXXXX.XXXXXXX}

\acmConference[xx]{xx}{xx}{xxx}
\acmPrice{xx}
\acmISBN{xxxx}

\usepackage{xspace}
\usepackage{balance}

\usepackage{pgfplots}
\usepackage{tikz}
\usepackage{multirow}
\usepackage{booktabs}
\usepackage{graphicx}
\usepackage{paralist}
\usepackage{array}
\usepackage{adjustbox}
\usepackage{caption}
\usepackage{todonotes}
\usepackage[caption=false]{subfig}
\newcolumntype{L}[1]{>{\raggedright\let\newline\\\arraybackslash\hspace{0pt}}m{#1}}
\usepackage{makecell}

\newcommand{\puball}{341\xspace} 
\newcommand{\pubexp}{324\xspace} 
\newcommand{\pubbench}{308\xspace} 

\newcommand{\pubpre}{123\xspace} 
\newcommand{\pubin}{212\xspace} 
\newcommand{\pubpost}{56\xspace} 
\newcommand{\pubmulti}{70\xspace} 
\newcommand{\benchunique}{137\xspace} 
\newcommand{\benchother}{51\xspace}
\newcommand{\metricsunique}{109\xspace} 

\newcommand{\setsplit}{232\xspace}
\newcommand{\setuprun}{143\xspace} 
\newcommand{\dataunique}{81\xspace} 
\newcommand{\dataavg}{2.7\xspace} 

\begin{document}

\title{Bias Mitigation for Machine Learning Classifiers: A Comprehensive Survey}


\author{Max Hort}
\affiliation{%
  \institution{Simula Research Laboratory}
  \city{Oslo}
  \country{Norway}}
\email{maxh@simula.no}

\author{Zhenpeng Chen}
\affiliation{%
  \institution{University College London}
  \city{London}
  \country{United Kingdom}}
\email{zp.chen@ucl.ac.uk}

\author{Jie M. Zhang}
\affiliation{%
  \institution{King's College London}
  \city{London}
  \country{United Kingdom}}
\email{jie.zhang@kcl.ac.uk}

\author{Mark Harman}
\affiliation{%
  \institution{University College London}
  \city{London}
  \country{United Kingdom}}
\email{mark.harman@ucl.ac.uk}

\author{Federica Sarro}
\affiliation{%
  \institution{University College London}
  \city{London}
  \country{United Kingdom}}
\email{f.sarro@ucl.ac.uk}

\renewcommand{\shortauthors}{Max Hort et al.}

\begin{abstract}
This paper provides a comprehensive survey of bias mitigation methods for achieving fairness in Machine Learning (ML) models. 
We collect a total of \puball publications concerning bias mitigation for ML classifiers. 
These methods can be distinguished based on their intervention procedure (i.e., pre-processing, in-processing, post-processing) and the technique they apply.
We investigate how existing bias mitigation methods are evaluated in the literature. 
In particular, we consider datasets, metrics and benchmarking.
Based on the gathered insights (e.g., What is the most popular fairness metric? How many datasets are used for evaluating bias mitigation methods?), we hope to support practitioners in making informed choices when developing and evaluating new bias mitigation methods.
\end{abstract}




\maketitle

\section{Introduction}\label{section_intro}
Machine Learning (ML) has been increasingly popular in recent years, both in the diversity and importance of applications~\citep{chouldechova2018frontiers}.
ML is used in a variety of critical applications such as justice risk assessments~\citep{angwin2016machine,berk2018fairness}, job recommendations~\citep{zhao2018learning}, and autonomous driving~\citep{corrabs230802935}.

While ML systems have the advantage to relieve humans from tedious tasks and are able to perform complex calculations at a higher speed~\citep{pessach2020algorithmic}, they are only as good as the data on which they are trained~\citep{barocas2016big}.
ML algorithms, which are never designed to intentionally incorporate bias, run the risk of replicating or even amplifying bias present in real-world data~\citep{pedreshi2008discrimination,barocas2016big,van2022overcoming}. This may cause unfair treatment in which some individuals or groups of people are \textit{privileged} (i.e., receive a favourable treatment) and others are \textit{unprivileged} (i.e., receive an unfavourable treatment).
In this context, a fair treatment of individuals constitutes that decisions are made independent of sensitive attributes such as gender or race, such that individuals are treated based on merit~\citep{kamiran2012decision,kamiran2018exploiting,mehrabi2021survey}.
For example, one can aim for an equal probability of population groups to receive a positive treatment, or an equal treatment of individuals that only differ in sensitive attributes.

Human bias has been transferred to various real-word systems relying on ML and there are many examples of this in the literature. For instance, bias has been found in advertisement and recruitment processes~\citep{zhao2018learning,jeffreydastin2018}, affecting university admissions~\citep{bickel1975sex} and human rights~\citep{mehrabi2021survey}.
Not only is such a biased behaviour undesired, but it can fall under regulatory control and risk the violation of anti-discrimination laws~\citep{pedreshi2008discrimination,chen2019fairness,romei2011multidisciplinary}, as sensitive attributes such as age, disability, gender identity, race are protected by US law in the Fair Housing Act and Equal Credit Opportunity Act~\citep{kuhlman2020no}.

Another example for a biased treatment of population groups can be found in the \textbf{COMPAS} (Correctional Offender Management Profiling for Alternative
Sanctions) software, used by courts in US to determine the risks of an individual to reoffend.
These scores are used to motivate decisions on whether and when defendants are to be set free, in different stages of the justice system.
Problematically, this software falsely labelled non-white defendants with higher risk scores than white defendants~\citep{angwin2016machine}.

To reduce the degree of bias that such systems exhibit, practitioners use three types of bias mitigation methods~\citep{friedler2019comparative}: 
\begin{itemize}
    \item \textbf{Pre-processing:}  bias mitigation in the training data, to prevent it from reaching ML models;
    \item \textbf{In-processing:} bias mitigation while training ML models;
    \item \textbf{Post-processing:} bias mitigation on trained ML models. 
\end{itemize}
In this survey, we use the terms ``bias mitigation'' and ``fairness improvement'' interchangeably and treat fairness as the absence of bias.

There has been a growing interest in fairness research, including definitions, measurements, and improvements of ML models~\citep{dunkelau2019fairness,pessach2020algorithmic,chouldechova2018frontiers,chen2022comprehensive,corrabs230801923}.
In particular, a variety of recent work addresses the mitigation of bias in binary classification models: given a collection of observations (training data) are labelled with a binary label (testing data)~\citep{verma2018fairness}. 

Despite the large amount of existing bias mitigation methods and surveys on fairness research, as \citet{pessach2020algorithmic} pointed out, there remain open challenges that practitioners face when designing new bias mitigation methods: ``It is not clear how newly proposed mechanisms should be
evaluated, and in particular which measures should be considered? which datasets should be used?
and which mechanisms should be used for comparison?''~\citep{pessach2020algorithmic}

To combat this challenge, we set out to perform a comprehensive survey of existing research on bias mitigation for ML models. 
We analyse \puball publications to identify practices applied in fairness research when creating bias mitigation methods.
In particular, we consider the datasets to which bias mitigation methods are applied, the metrics used to determine the degree of bias, and the approaches used for benchmarking the effectiveness of bias mitigation methods.
By doing so, we allow practitioners to focus their effort on creating bias mitigation methods rather than requiring a lot of time to determine their experimental setup (e.g., which datasets to test on, which benchmark to consider).

To the best of our knowledge, this is the most comprehensive survey to systematically search and cover bias mitigation methods and their empirical evaluation.
To summarize, the contributions of this survey are:
\begin{enumerate}
    \item we provide a comprehensive overview of the research on bias mitigation methods for ML classifiers; 
     \item we introduce the experimental design details for evaluating existing bias mitigation methods;
    \item we identify challenges and opportunities for future research on bias mitigation methods.
    \item we make the collected paper repository public, to allow for future replication and manual investigation of our results: \url{https://solar.cs.ucl.ac.uk/os/softwarefairness.html}.
\end{enumerate}

The rest of this paper is structured as follows.
Section \ref{section_background} presents an overview of related surveys.
The search methodology is described in Section \ref{section_method}.
Sections \ref{section:algs}-\ref{section:benchmarking} describe research on bias mitigation methods. 
Opportunities and challenges that the field of fairness research and bias mitigation methods face are discussed in Section \ref{section:discussion}.
Section \ref{section:recommendations} provides recommendations to practitioners, distilled from the collected publications. 
Section \ref{section_conclusion} concludes this survey.

\section{Related Surveys}\label{section_background}
In this section, we provide an overview of existing surveys in the fairness literature and their contents.
This allows us to identify the knowledge gap filled by our survey.

\citet{mehrabi2021survey} and \citet{pessach2020algorithmic} provided an overview of bias and discrimination types, fairness definitions and metrics, bias mitigation methods, and existing datasets.
For example, \citet{pessach2022review} listed the datasets and metrics used by 27 bias mitigation methods.
A similar focus has been pursued by \citet{dunkelau2019fairness}, who provided an extensive overview on fairness notions, available frameworks, and bias mitigation methods for classification problems.
They moreover provided a classification of approaches for each type (i.e., pre-, in-, and post-processing). 
The most exhaustive categorization of bias mitigation methods, to date, has been conducted by \citet{caton2020fairness}, who also presented fairness metrics and fairness platforms.

A detailed collection of prominent fairness definitions for classification problems is provided by \citet{verma2018fairness}.
Similarly, \citet{zliobaite2015survey} surveyed measures for indirect discrimination for ML. 
While these collections describe current metrics used to determine the fairness of ML models, \citet{hutchinson201950} drew parallels from fairness research in the 1960s and 1970s concerning test fairness, for education and hiring, to current advances.
Similar to modern metrics and evaluation approaches, past work considered fairness with regards to individuals and groups, or the use of confusion matrix measures (Section~\ref{section:metrics}).

In addition to the surveys on fairness metrics, \citet{le2022survey} provided a survey with 15 frequently used datasets in fairness research.
For each dataset, they described the available features and their relationships with sensitive attributes.

Other surveys are concerned with fairness and consider the following perspectives: learning-based sequential decision algorithms~\citep{zhang2021fairness}, criminal justice~\citep{berk2018fairness}, graph representations~\citep{zhang2022fairness}, ML testing~\citep{zhang2020machine}, Software Engineering~\citep{soremekun2022software,chen2022fairness}, or Natural Language Processing~\citep{sun2019mitigating,blodgett2020language}.

While previous surveys focused on ML classification, and some mentioned bias mitigation methods, none has yet systematically covered the evaluation bias mitigation methods (e.g., how are methods benchmarked, what dataset are used).
The surveys related closest to ours are provided by \citet{dunkelau2019fairness}, and \citet{pessach2022review}. 

\citet{dunkelau2019fairness} provided an overview of bias mitigation methods with a focus on their implementation and underlying algorithms. However, further evaluation details of these methods, such as dataset and metric usage, were not addressed.
While \citet{pessach2022review} listed the datasets and metrics used by 27 bias mitigation methods, they did not provide actionable insights to support developers.
In addition to combining aspects of both surveys (i.e., extensive collection of bias mitigation methods like \citet{dunkelau2019fairness}, and providing information on datasets and metrics similar to \citet{pessach2020algorithmic}), we aim to analyze the findings of a comprehensive literature search to devise recommendations.


\section{Survey Methodology}\label{section_method}

The purpose of this survey is to gather and categorize research work that mitigates bias in ML models.
Given that the existing literature focuses on classification for tabular data, this survey also focuses on bias mitigation methods for such classification tasks.




\subsection{Search Methodology}\label{method_search}
This section outlines our search procedure. We start with a preliminary search, followed by a repository search and snowballing.

\vspace{2mm}
\noindent{\textbf{Preliminary Search.}}
Prior to systematically searching online repositories, we conduct a preliminary search.
The goal of the preliminary search is to gain a deeper understanding of the field and assess whether there is a sufficient number of publications to allow for subsequent analysis.
In particular, we collect bias mitigation publications from four existing surveys (see Section \ref{section_background}):
\begin{itemize}
    \item \citet{mehrabi2021survey} : 24 bias mitigation methods;
    \item \citet{pessach2022review}: 30 bias mitigation methods;
    \item \citet{dunkelau2019fairness}: 40 bias mitigation methods;
    \item \citet{caton2020fairness}: 70 bias mitigation methods.
\end{itemize}
In total, we collect 100 unique publications with bias mitigation methods from these four surveys.

\vspace{2mm}
\noindent{\textbf{Repository Search.}}\label{repo}
After the preliminary search, we conduct a search of six established online repositories (IEEE, ACM, ScienceDirect, Scopus, arXiv, and Google Scholar).

The search procedure is guided by two groups of keywords:
\begin{itemize}
    \item Domain: machine learning, deep learning, artificial intelligence;
    \item Bias Mitigation: fairness-aware, discrimination-aware, bias mitigation, debias*, unbias*;
\end{itemize}
In this context, \textit{Domain} keywords ensure that the bias discussed in the publication affects machine learning systems. \textit{Bias Mitigation} keywords ensure that the publication addresses bias reduction via the use of bias mitigation methods.
For the six repositories, we collected publications that contain at least one \textit{Domain} and one \textit{Bias mitigation} keyword (i.e., we check each possible combination of keywords for the two categories).

\vspace{2mm}
\noindent{\textbf{Selection.}}\label{method_selection}
To ensure that the publications included in this survey are relevant to the context of bias mitigation for ML models, we consider the following {\bf inclusion criteria}: 1) describe human biases; 2) address classification problems; 3) use tabular data (e.g., do not make decisions based on images or text alone).

To ensure that irrelevant publications are excluded from the search results, we manually check publications in three stages~\citep{martin2016survey}:
\begin{enumerate}
    \item \textbf{Title:} Publications with irrelevant titles to the survey are excluded;
    \item \textbf{Abstract:} The abstract of every publication is checked. Publications that show to be irrelevant to the survey at this step are excluded (e.g. not about ML, do not apply debiasing);
    \item \textbf{Body:} For publications that passed the previous two steps, we check the entire publication to determine whether they satisfy the inclusion criteria. If not, they are excluded.
\end{enumerate}

\vspace{2mm}
\noindent{\textbf{Snowballing.}}
\label{sec:snow}
After conducting the repository search, we apply backward snowballing (i.e., finding new publications that are cited by publications we already selected) for each publication retained after the ``Body'' stage~\citep{wohlin2014guidelines}.
This snowballing step is repeated for every new publication found.
The goal of snowballing is to find missing related work with regards to the collected publications.
This is in particular useful if undiscovered bias mitigation methods are used for benchmarking. 

\begin{table}[]
\caption{Publications found at each stage of the search procedure.}
\centering
\label{table:search-results}
\begin{tabular}{lr}
\toprule
Stage                    & Publications \\
\midrule
Preliminary search       & 100          \\
Repository search Oct'21 & 75           \\
Repository search Jul'22 & 56           \\
Snowballing              & 78           \\ 
Author feedback          & 32        \\\midrule
Total                    & \puball      \\ \bottomrule
\end{tabular}
\end{table}

\subsection{Selected Publications}
In total, we gathered \puball publications over the different stages of our search procedure.
Table \ref{table:repo} summarises the results of the two repository searches. The first search was conducted from the 7th of October to 10th of October 2021, and the second search was conducted on the 21st of July 2022. 
The purpose of the second search is to collect publications from the year 2022 (i.e., we filtered search results for the publication year 2022). 
In October 2021, Google Scholar provided $8,738$ publications that were in line with the search keywords. We restricted our search to the first $1,000$ entries as prioritised by Google Scholar based on relevance.
Similarly, the second search yielded $1,995$ results and we focused on the first $1,000$ entries.

To ensure that our survey is comprehensive and accurate, we contacted the corresponding authors of the 309 publications collected via the preliminary search, the two repository searches and snowballing. We asked them to check whether our description about their work is correct. 
Based on their feedback, we included additional 32 publications.
In Table~\ref{table:search-results} we summarise the number of publications we found at each step of the search.

\begin{table}[!htb]
\caption{Results of the repository search. For each of the six repositories, we show the number of publications retained after each filtering stage. The ``Body'' column shows the number of publications included in this survey.}
\centering
\label{table:repo}
\begin{minipage}{\columnwidth}
    \centering
    \begin{tabular}{lrrrr}
        \toprule
        Repository    & Initial & Title & Abstract & Body \\ \midrule
        ACM           & 118   & 26    & 16       & 13    \\ 
        ScienceDirect & 166   & 9     & 5        &  3    \\
        IEEE          & 401   & 18    & 9        &  9    \\
        arXiv         & 650   & 69    & 48       & 38    \\
        Scopus        & 1063  & 44    & 28       & 21    \\
        Google Scholar       & 8738  & 119   & 90       & 77    \\
        \bottomrule
    \end{tabular}
    \caption*{Search results October'21.}
\end{minipage}\hfill 

\begin{minipage}{\columnwidth}
    \centering
    \begin{tabular}{lrrrr}
        \toprule
        Repository    & Initial & Title & Abstract & Body \\ \midrule
        ACM           & 468   & 17    & 14       & 8    \\ 
        ScienceDirect & 88   & 6     & 3        &  2    \\
        IEEE          & 90   & 8    & 1        &   1    \\
        arXiv         & 465   & 42    & 23       &  17   \\
        Scopus        & 356  & 13    & 9       &  5   \\
        Google Scholar & 1995  & 62   & 51       &  35   \\
        \bottomrule
    \end{tabular}
    \caption*{Search results July'22.}
  \end{minipage}
\end{table}

\begin{figure*}[]
\begin{minipage}{\linewidth}
 \subfloat[Pre-processing.]{%
  \includegraphics[width=0.32\linewidth]{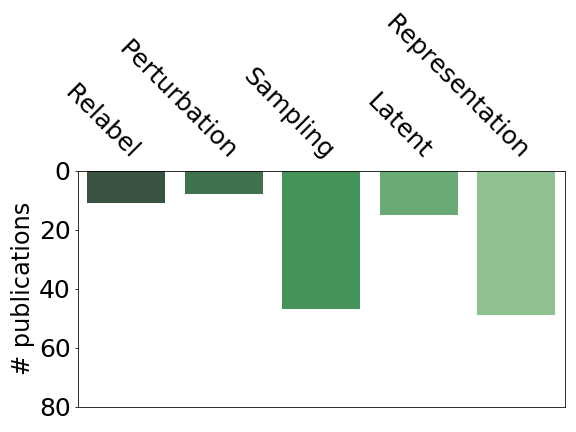}
}
  \subfloat[In-processing.]{%
 \includegraphics[width=0.32\linewidth]{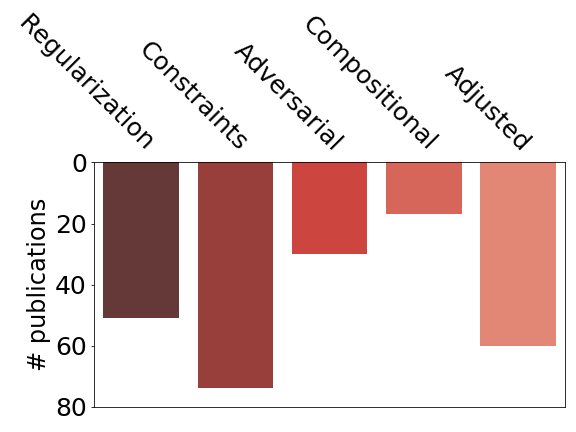}
 }
  \subfloat[Post-processing.]{%
 \includegraphics[width=0.32\linewidth]{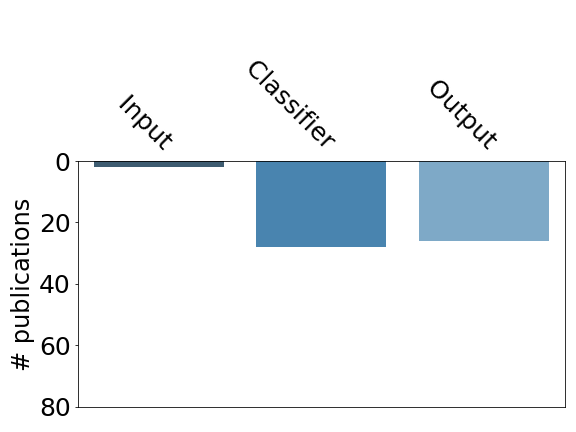}
 }
\end{minipage}

\caption{Categorization of bias mitigation methods. Categories are grouped based on their type (i.e., pre-processing, in-processing, post-processing) and the number of publications of each category is shown.}
\label{fig:approaches}
\end{figure*}

\begin{figure}
\centering
  \includegraphics[width=.8\columnwidth]{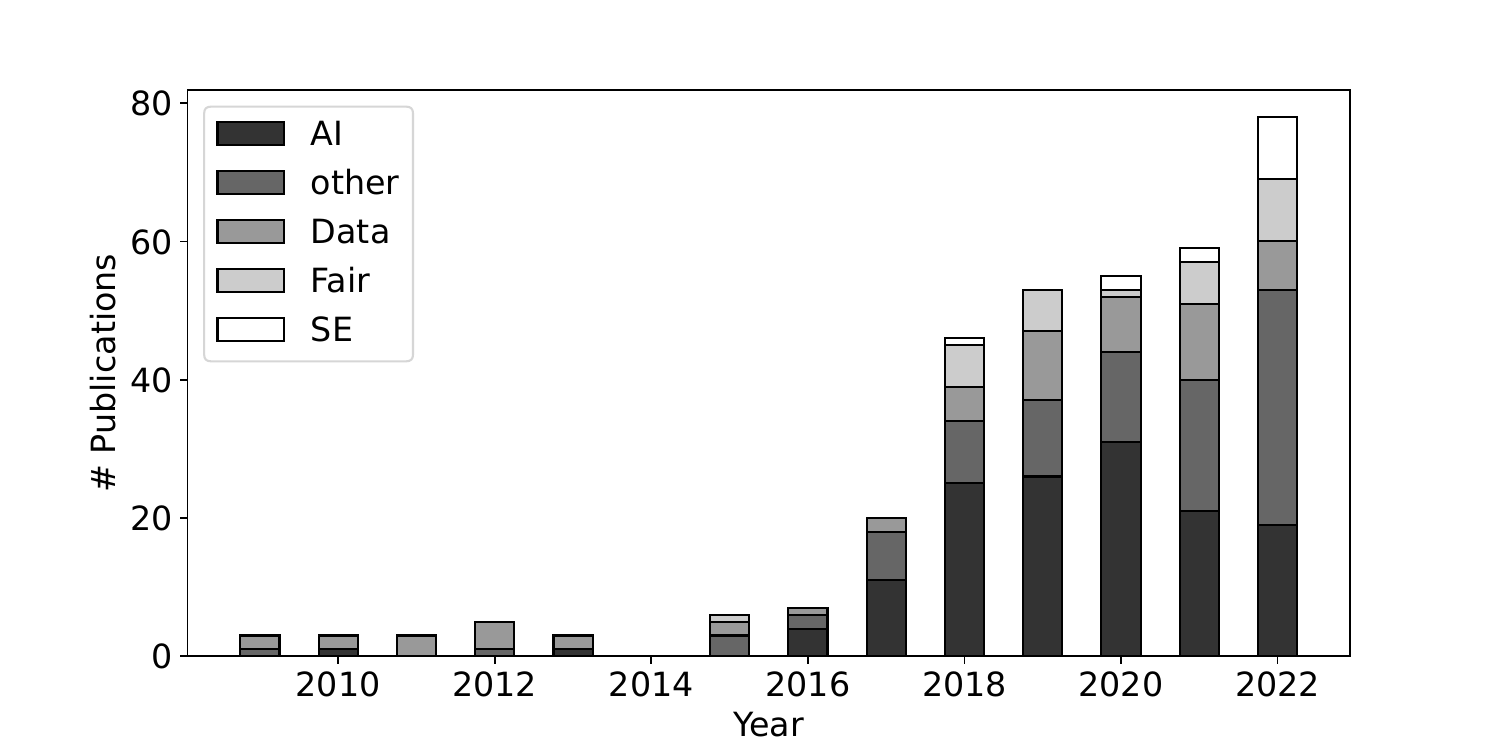}
  \caption{Number of publications per year and venue type.}
  \label{fig:year}
\end{figure}

In Figure \ref{fig:year} we show the distribution of the publications per year and venue type.
We categorized the \puball publications in five venue types, in line with the categories by~\citet{soremekun2022software}: Artificial Intelligence (AI), Data, Fairness, Software Engineering (SE), other.
Note that the category ``other'' consists of 100 publications, 68 of which are published on arXiv. The category SE combines publications form Software Engineering, Programming Language and Security venues.
From this figure, we can see that there is an increasing interest in bias mitigation methods and a steady increase of publications over the years.
In particular, we observe a huge jump in the number of publications in 2018, more than doubling the number of publications from 2017 (i.e., from 20 to 46 publications).
Prior years, from 2009-2016, have seen less than 10 publications each.
The venues with the highest number of 
publications are: NeurIPS (38 publications), ICML (27 publications), AAAI (18 publications), FAccT (13 publications), AIES (12 publications).

\subsection{Visibility}
In this section, we address the visibility of bias mitigation methods by using the amount of publications and number of citations as a proxy for bias mitigation method visibility across different venues.\footnote{We obtained the number of citations for each publication from Google Scholar on February 24th, 2023, and included them in our online repository~\citep{homepage}.}

As shown in Figure~\ref{fig:year}, there is an increasing trend in the number of publications on bias mitigation methods per year, which supports the claim that the visibility and relevance of bias mitigation is growing. 
Among the five venue types (AI, Data, Fairness, SE, other), bias mitigation methods exhibit the highest visibility in terms of number of publications for AI (139 publications), data (59 publications) and other venues (most notably arXiv, with 68 publications).
The past five years, from 2018 onwards, saw an uptake of bias mitigation methods in a wider range of venues, with the inclusion of bias mitigation methods in Software Engineering venues and the creation of the ACM Conference on Fairness, Accountability, and Transparency (ACM FAccT),\footnote{\url{https://facctconference.org/index.html}} as well as specialised venues co-located with well-renowned international conference such as the IEEE/ACM International Workshop on Equitable Data \& Technology (FairWare) at the International Conference of Software Engineering.\footnote{\url{https://dblp.org/db/conf/fairware-ws/index.html}}

\begin{figure}
\centering
\includegraphics[width=0.65\columnwidth]{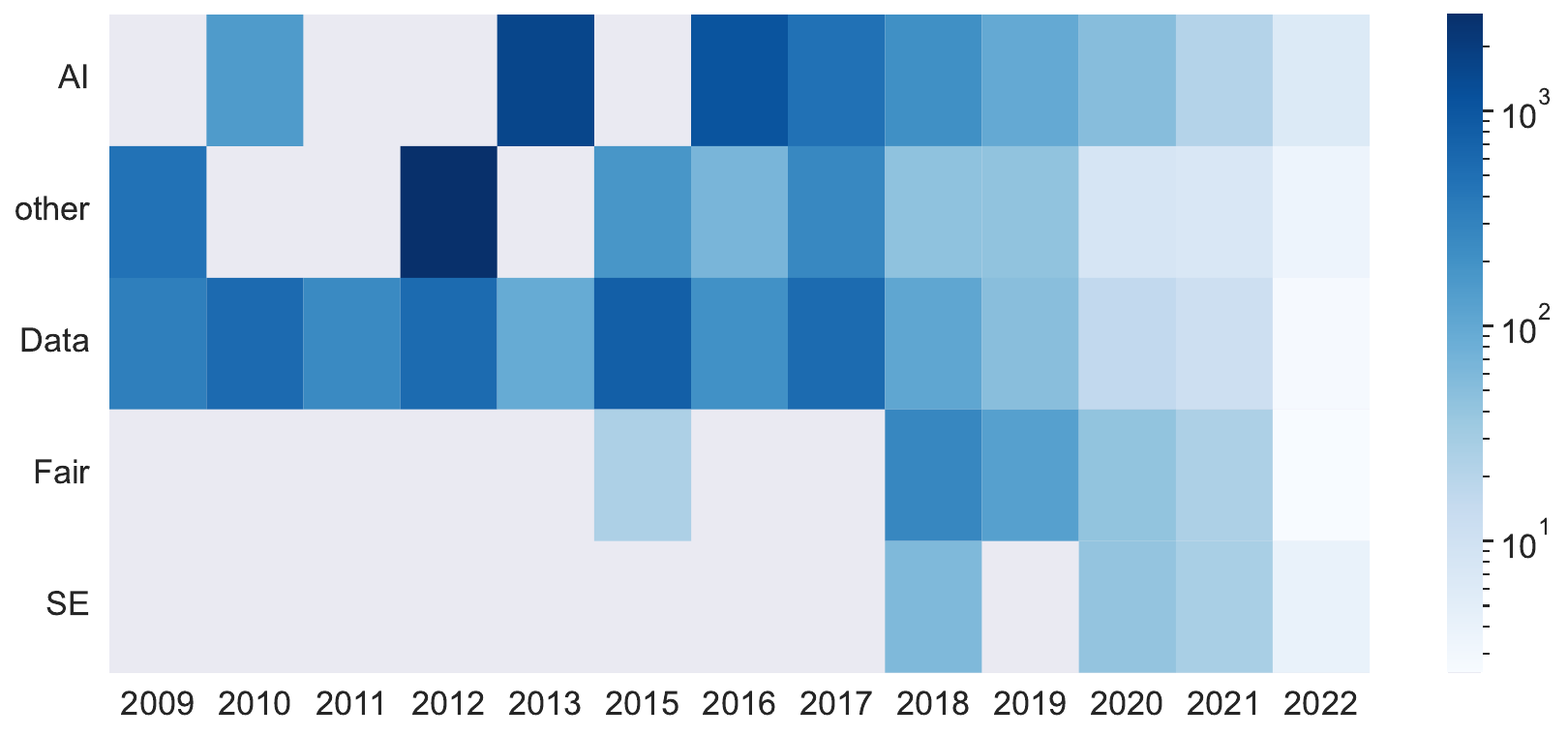}
  \caption{Average number of citations per year and venue type.}
  \label{fig:citesHeatmap}
\end{figure}

Figure~\ref{fig:citesHeatmap} provides a closer look at the average number of citations of publications per venue type.
We can observe that publications from early years of bias mitigation methods have a high average visibility (i.e., number of citations). 
A reason for this can be found in the low number of publications, with only 3-7 publications yearly from 2009-2016, and the relevance of such publications to be the foundation of proceeding work. 
Data venues published bias mitigation methods consistently, every year, from 2009 to 2022.
While  Fairness and SE venues have fewer publications per year, the respective papers achieve a high visibility, frequently with a higher average number of citations than Data and AI venues for the same years.
The highest average number of citations was achieved by publications in fairness venues in 2018, 2019 and 2021.

Among the most cited publications (19 of which publications have been cited more than 500 times) only two have not been published in AI or data venues.
This includes the work by \citet{dwork2012fairness} (published in the proceedings of the 3rd innovations in theoretical computer science conference) and \citet{zhang2018mitigating}	(published at the AAAI/ACM Conference on AI, Ethics, and Society).
We note that 10 out of the 15 most cited works have publicly available implementations in fairness frameworks~\citep{bantilan2018themis,bird2020fairlearn,bellamy2018ai}.

\newpage
\subsection{Limitations}
This survey focuses on investigating the fairness of ML models from an algorithmic point of view.
While fairness is a multi-disciplinary field of research, and has been addressed by various communities, including law~\citep{blumrosen1978wage}, health studies~\citep{obermeyer2019dissecting}, and criminal justice~\citep{bao2021s}, we focus on the algorithmic fairness and bias as exhibited by ML models.

Moreover, our search procedure is designed to find publications that mitigate bias for tabular data.
This does not mean that we exclude a priori relevant publications if they have been published at Computer Vision or Natural Language Processing venues.
In fact, such publications are considered in our survey if bias mitigation for tabular data is addressed, whereas bias mitigation methods that are solely applied for visual or textual tasks are not included.

Furthermore, we note that the overview presented herein is based on bias mitigation methods as proposed by the research community, often applied to publicly available data.  While these dataa can be based on real-world scenarios, results might not transfer to real-world applications~\citep{bao2021s, ding2021retiring}.

\section{Algorithms}
\label{section:algs}
In this section, we discuss the bias mitigation methods found in our literature search.
We distinguished bias mitigation methods based on their type (i.e., in which stage of the ML process are they applied): pre-processing (Section \ref{section:pre}), in-processing (Section \ref{section:in}) and post-processing (Section \ref{section:post}) methods~\citep{friedler2019comparative}.  
Moreover, we organize methods in categories (i.e., the bias mitigation approach).  For this, we follow taxonomies devised by \citet{dunkelau2019fairness}, as well as \citet{caton2020fairness}.
Figure~\ref{fig:approaches} illustrates the 13 categories we use.

Among the \puball publications, \pubpre used pre-processing (Section \ref{section:pre}), \pubin used in-processing (Section \ref{section:in}) and \pubpost used post-processing methods (Section \ref{section:post}).
We observe that a single publication may investigate up to three different types of bias mitigation methods, and as such it can be counted multiple times (for example, an approach can apply pre-processing before adapting the training procedure during an in-processing stage).  This is the case for \pubmulti publications analysed in this survey, for which we provide more information in Section \ref{section:combined}.

\subsection{Pre-processing Bias Mitigation Methods}
\label{section:pre}
In this section, we present bias mitigation methods that combat bias by applying changes to the training data.
Table~\ref{table:pre} lists the \pubpre publications we found, according to the type of pre-processing method used. 

\begin{table}[]
\centering
\caption{Publications on Pre-processing bias mitigation methods.}
\label{table:pre}
\begin{adjustbox}{max width=0.42\columnwidth}
\begin{tabular}[t]{llll}
\toprule
Type & Authors [Ref] & Year & Venue \\ \midrule

\multirow{11}{*}{\rotatebox{90}{Relabel}} & \citet{calders2009building} & 2009 & ICDMW \\
 & \citet{kamiran2009classifying} & 2009 & ICCCC \\
 & \citet{vzliobaite2011handling} & 2011 & ICDM \\
 & \citet{luong2011k} & 2011 & KDD \\
 & \citet{hajian2012methodology} & 2012 & TKDE \\
 & \citet{kamiran2012data} & 2012 & KAIS \\
 & \citet{zhang2018achieving} & 2018 & IJCAI \\
 & \citet{iosifidis2019fairness} & 2019 & DEXA \\
 & \citet{seker2022preprocessing} & 2022 & HTI \\
 & \citet{sun2022towards} & 2022 & EuroS\&P \\
 & \citet{alabdulmohsin2022reduction} & 2022 & arXiv \\ \midrule
\multirow{8}{*}{\rotatebox{90}{Perturbation}} & \citet{hajian2012methodology} & 2012 & TKDE \\
 & \citet{feldman2015certifying} & 2015 & KDD \\
 & \citet{lum2016statistical} & 2016 & arXiv \\
 & \citet{wang2018avoiding} & 2018 & NeurIPS \\
 & \citet{johndrow2019algorithm} & 2019 & Ann Appl Stat \\
 & \citet{wang2019repairing} & 2019 & ICML \\
 & \citet{li2022propose} & 2022 & SSRN \\
 & \citet{li2022training} & 2022 & ICSE \\ \midrule
\multirow{47}{*}{\rotatebox{90}{Sampling}} & Calders et al.~\citet{calders2009building} & 2009 & ICDMW \\
 & \citet{kamiran2010classification} & 2010 & BNAIC \\
 & \citet{vzliobaite2011handling} & 2011 & ICDM \\
 & \citet{kamiran2012data} & 2012 & KAIS \\
 & \citet{zhang2017causal} & 2017 & IJCAI \\
 & \citet{chen2018my} & 2018 & NeurIPS \\
 & \citet{iosifidis2018dealing} & 2018 & report \\
 & \citet{xu2018fairgan} & 2018 & Big Data \\
 & \citet{krasanakis2018adaptive} & 2018 & TheWebConf \\
 & \citet{abusitta2019generative} & 2019 & arXiv \\
 & \citet{xu2019achievingan} & 2019 & IJCAI \\
 & \citet{zelaya2019parametrised} & 2019 & KDD \\
 & \citet{salimi2019interventional} & 2019 & MOD \\
 & \citet{iosifidis2019fairness} & 2019 & DEXA \\
 & \citet{iosifidis2019fae} & 2019 & Big Data \\
 & \citet{xu2019fairgan} & 2019 & Big Data \\
 & \citet{abay2020mitigating} & 2020 & arXiv \\
 & \citet{hu2020fairnn} & 2020 & DS \\
 & \citet{chakraborty2020fairway} & 2020 & FSE \\
 & \citet{jiang2020identifying} & 2020 & AISTATS \\
 & \citet{sharma2020data} & 2020 & AIES \\
 & \citet{celis2020data} & 2020 & ICML \\
 & \citet{morano2020bias} & 2020 & Thesis \\
 & \citet{yan2020fair} & 2020 & CIKM \\
 & \citet{chuang2021fair} & 2021 & ICLR \\
 & \citet{salazar2021fawos} & 2021 & IEEE Access \\
 & \citet{zhang2021farf} & 2021 & PAKDD \\
 & \citet{yu2021fair} & 2021 & arXiv \\
 & \citet{iofinova2021flea} & 2021 & arXiv \\
 & \citet{roh2021sample} & 2021 & NeurIPS \\
 & \citet{du2021robust} & 2021 & CIKM \\
 & \citet{singh2022developing} & 2021 & MAKE \\
 & \citet{amend2021improving} & 2021 & JCSC \\
 & \citet{jang2021constructing} & 2021 & AAAI \\
 & \citet{verma2021removing} & 2021 & arXiv \\
 & \citet{fairsmote} & 2021 & FSE \\
 & \citet{cruz2021bandit} & 2021 & ICDM \\
 & \citet{pmlr-v162-wang22ac} & 2022 & ICML \\
 & \citet{pentyala2022privfairfl} & 2022 & arXiv \\
 & \citet{rajabi2022tabfairgan} & 2022 & MAKE \\
 & \citet{sun2022towards} & 2022 & EuroS\&P \\
 & \citet{dablain2022towards} & 2022 & arXiv \\
 & \citet{zhenpengmaat2022} & 2022 & FSE \\
 & \citet{li2022achieving} & 2022 & PMLR \\
 & \citet{chakraborty2022fair} & 2022 & FairWARE \\
 & \citet{Almuzaini2022abc} & 2022 & FAccT \\
 & \citet{chai2022fairness} & 2022 & ICML \\
 
\bottomrule 
\end{tabular}
\end{adjustbox}
\begin{adjustbox}{max width=0.42\columnwidth}
\begin{tabular}[t]{llll}
\toprule
Type & Authors [Ref] & Year & Venue \\ \midrule
\multirow{15}{*}{\rotatebox{90}{Latent}} & \citet{calders2010three} & 2010 & DMKD \\
 & \citet{kilbertus2017avoiding} & 2017 & NeurIPS \\
 & \citet{gupta2018proxy} & 2018 & arXiv \\
 & \citet{madras2019fairness} & 2019 & FAccT \\
 & \citet{oneto2019taking} & 2019 & AIES \\
 & \citet{wei2020optimized} & 2020 & PMLR \\
 & \citet{kehrenberg2020tuning} & 2020 & Front. Artif. Intell. \\
 & \citet{grari2021fairness} & 2021 & arXiv \\
 & \citet{chen2022fair} & 2022 & arXiv \\
 & \citet{liang2022joint} & 2022 & arXiv \\
 & \citet{jung2022learning} & 2022 & CVPR \\
 & \citet{Diana2022multi} & 2022 & FAccT \\
 & \citet{chakraborty2022fair} & 2022 & FairWARE \\
 & \citet{wu2022fair} & 2022 & CLeaR \\
 & \citet{suriyakumar2022personalization} & 2022 & arXiv \\ \midrule
\multirow{49}{*}{\rotatebox{90}{Representation}} & \citet{zemel2013learning} & 2013 & ICML \\
 & \citet{edwards2015censoring} & 2015 & arXiv \\
 & \citet{Louizos2016} & 2016 & ICLR \\
 & \citet{perez2017fair} & 2017 & ECML PKDD \\
 & \citet{calmon2017optimized} & 2017 & NeurIPS \\
 & \citet{hacker2017continuous} & 2017 & arXiv \\
 & \citet{komiyama2017two} & 2017 & arXiv \\
 & \citet{xie2017controllable} & 2017 & NeurIPS \\
 & \citet{mcnamara2017provably} & 2017 & arXiv \\
 & \citet{du2018data} & 2018 & IEEE J Sel \\
 & \citet{grgic2018beyond} & 2018 & AAAI \\
 & \citet{madras2018learning} & 2018 & ICML \\
 & \citet{samadi2018price} & 2018 & NeurIPS \\
 & \citet{quadrianto2018neural} & 2018 & arXiv \\
 & \citet{moyer2018invariant} & 2018 & NeurIPS \\
 & \citet{song2019learning} & 2019 & AISTATS \\
 & \citet{gordaliza2019obtaining} & 2019 & ICML \\
 & \citet{quadrianto2019discovering} & 2019 & CVPR \\
 & \citet{creager2019flexibly} & 2019 & ICML \\
 & \citet{wang2019approaching} & 2019 & arXiv \\
 & \citet{lahoti2019ifair} & 2019 & ICDE \\
 & \citet{feng2019learning} & 2019 & arXiv \\
 & \citet{lahoti2019operationalizing} & 2019 & VLDB \\
 & \citet{Zhao2020Conditional} & 2020 & ICLR \\
 & \citet{tan2020learning} & 2020 & AISTATS \\
 & \citet{jaiswal2020invariant} & 2020 & AAAI \\
 & \citet{zehlike2020matching} & 2020 & DMKD \\
 & \citet{sarhan2020fairness} & 2020 & ECCV \\
 & \citet{madhavan2020fairness} & 2020 & CIKM \\
 & \citet{kim2020fair} & 2020 & AAAI \\
 & \citet{ruoss2020learning} & 2020 & NeurIPS \\
 & \citet{fong2021fairness} & 2021 & arXiv \\
 & \citet{gupta2021controllable} & 2021 & AAAI \\
 & \citet{zhu2021learning} & 2021 & ICCV \\
 & \citet{grari2021learning} & 2021 & ECML PKDD \\
 & \citet{salazar2021automated} & 2021 & VLDB \\
 & \citet{oh2022learning} & 2022 & arXiv \\
 & \citet{Agarwal2022power} & 2022 & FAccT \\
 & \citet{wu2022semi} & 2022 & arXiv \\
 & \citet{shui2022fair} & 2022 & arXiv \\
 & \citet{qi2022fairvfl} & 2022 & arXiv \\
 & \citet{balunovic2022fair} & 2022 & ICLR \\
 & \citet{kairouz2022censored} & 2022 & T-IFS \\
 & \citet{liu2022fair} & 2022 & Neural Process. Lett. \\
 & \citet{cerrato2022fair} & 2022 & arXiv \\
 & \citet{kamani2022efficient} & 2022 & Mach. Learn. \\
 & \citet{rateike2022don} & 2022 & FAccT \\
 & \citet{Galhotra2022causal} & 2022 & SIGMOD \\
 & \citet{KIM202226} & 2022 & Neurocomputing \\ 
\bottomrule 
\end{tabular}
\end{adjustbox}
\end{table}
\subsubsection{Relabelling and Perturbation}
\label{section:pre-relabel}
This section presents bias mitigation methods that apply changes to the values of the training data.
Changes have been applied to the ground truth labels (relabelling) or the remaining features (perturbation).

A popular approach for relabelling datasets is ``massaging'', proposed by \citet{kamiran2009classifying}. 
In the first stage, ``massaging'' uses a ranker to determine the best candidates for relabelling.
In particular, instances close to the decision boundary are selected, to minimize the negative impact of relabelling on accuracy. 
Typically, an equal amount of instances with positive and negative labels are selected, according to their rank and their labels are switched.

Massaging has later been extended by \citet{kamiran2012data}, and \citet{calders2009building}.
Moreover, \citet{vzliobaite2011handling}
created a related method called ``local massaging''.
``Massaging'' has also been applied by other work~\citep{iosifidis2019fairness,zhang2018achieving}.

Another relabelling approach was proposed by \citet{luong2011k}, who relabelled instances based on their $k$-nearest neighbours, such that similar individuals receive similar labels.

\citet{feldman2015certifying} used perturbation to modify non-protected attributes, such that their values for privileged and unprivileged groups are comparable. In particular, the values are adjusted to bring their distributions closer together while preserving the respective ranks within a group (e.g., the highest values of attribute $a$ for the privileged group remains highest after perturbation).
\citet{lum2016statistical,johndrow2019algorithm} used conditional models for perturbation, which allowed for modification of multiple variables (continuous or discrete).
\citet{li2022propose} proposed an iterative approach for perturbation.
At each step, the most bias-prone attribute is selected and transformed, until the degree of bias exhibited by a classification model is below a specified threshold.


Other than perturbing the underlying data for all groups to move them closer~\citep{feldman2015certifying,lum2016statistical,johndrow2019algorithm},
\citet{wang2018avoiding,wang2019repairing} considered only the unprivileged group for perturbation, seeking to resolve disparity by improving the performance of the unprivileged group.
\citet{hajian2012methodology} applied both relabeling and perturbation (i.e., changes to the sensitive attribute).

\subsubsection{Sampling}
\label{section:pre-sampling}
Sampling methods change the training data by changing the distribution of samples (e.g., adding, removing samples) or adapting their impact on training. 
Similarly, the impact of training data instances can be adjusted by reweighing their importance~\citep{calders2009building,kamiran2012data,iosifidis2019fairness,celis2020data,du2021robust,yu2021fair,li2022achieving,Almuzaini2022abc,chai2022fairness,pentyala2022privfairfl,abay2020mitigating}.

Reweighing was first introduced by \citet{calders2009building}.
Each instance receives a weight according to its label and protected attribute (e.g., instances in the unprivileged group and positive label receive a higher weight as this is less likely). In the training process of classification models, a higher instance weight causes higher losses when misclassified.
Weighted instances are sampled with replacement according to their weights.
If the classification model is able to process weighted instances, the dataset can be used for training without resampling~\citep{kamiran2012data}.

\citet{jiang2020identifying} and \citet{krasanakis2018adaptive} used reweighing to combat biased labels in the original training data.

Instead of assigning equal weights to data instances of the same population subgroup, \citet{li2022achieving} assigned individual weights to instances of the training data.

Other sampling strategies include the removal of data points (downsampling)~\citep{chakraborty2020fairway,verma2021removing,roh2021sample,salimi2019interventional,cruz2021bandit,zhang2021farf,zhenpengmaat2022,iofinova2021flea,pmlr-v162-wang22ac} or the addition of new data points (upsampling).
Popular methods for upsamplig are oversampling for duplicating instances of the minority group~\citep{iosifidis2018dealing,zelaya2019parametrised,morano2020bias,amend2021improving} and the use of SMOTE~\citep{chawla2002smote}. SMOTE does not duplicate instances but generates synthetic ones in the neighborhood of the minority group~\citep{iosifidis2018dealing,yan2020fair,fairsmote,zelaya2019parametrised,salazar2021fawos,morano2020bias,chakraborty2022fair,singh2022developing,dablain2022towards}.

To sample datapoints, uniform~\citep{kamiran2012data} and preferential~\citep{kamiran2012data,kamiran2010classification,vzliobaite2011handling,hu2020fairnn,zelaya2019parametrised} strategies have been followed, where preferential sampling changes the distribution of instances close to the decision boundary. 

\citet{xu2018fairgan,xu2019achievingan,xu2019fairgan} used a generative approach to generate discrimination-free data for training~\citep{abusitta2019generative,rajabi2022tabfairgan,jang2021constructing}.
\citet{zhang2017causal} used causal networks to create a new dataset. The initial dataset is used to create a causal network, which is then modified to reduce discrimination. The debiased causal network is used to generate a new dataset.
\citet{sharma2020data} created additional data for augmentation by duplicating existing datasets and swapping the protected attribute of each instance. The newly-created data is successively added to the existing dataset.

\subsubsection{Latent variables}
\label{section:pre-latent}
Latent variables describe the augmentation of training data with additional features that are preferably unbiased.
In previous work, latent variables have been used to represent labels~\citep{wei2020optimized,kehrenberg2020tuning} and group memberships (i.e., protected or unprotected group)~\citep{gupta2018proxy,grari2021fairness,oneto2019taking,chakraborty2022fair,liang2022joint,jung2022learning,Diana2022multi,chen2022fair,suriyakumar2022personalization}, and are  frequently considered when dealing with causal graphs~\citep{madras2019fairness,grari2021fairness,kilbertus2017avoiding}. 

For instance, \citet{calders2010three} clustered the instances to detect those that should receive a positive latent label and those that should receive a negative one.
For this purpose, they used an expectation maximization algorithm. 


\citet{gupta2018proxy} tackled the problem of bias mitigation for situations where group labels are missing in the datasets. To combat this issue, they created a latent ``proxy'' variable for the group membership and incorporated constraints for achieving fairness for such proxy groups in the training procedure.



\subsubsection{Representation}
\label{section:pre-representation}
\textit{Representation} learning aims at learning a transformation of the training data such that bias is reduced while maintaining as much information as possible.

The first representation learning approach for bias mitigation was Learning Fair Representations (LFR), proposed by \citet{zemel2013learning}.
LFR translates representation learning into an optimization problem with two objectives: 1) removing information about the protected attribute; 2) minimizing the information loss of non-sensitive attributes.

A popular used approach for generating fair representations is optimization~\citep{calmon2017optimized,moyer2018invariant,lahoti2019ifair,gordaliza2019obtaining,lahoti2019operationalizing,mcnamara2017provably,zehlike2020matching,hacker2017continuous,du2018data,song2019learning,shui2022fair}.
Other used techniques are: 
\begin{itemize}
    \item adversarial learning~\citep{edwards2015censoring,madras2018learning,xie2017controllable,feng2019learning,ruoss2020learning,jaiswal2020invariant,kairouz2022censored,kim2020fair,grari2021learning,zhu2021learning,Zhao2020Conditional,qi2022fairvfl};
    \item variational autoencoders~\citep{Louizos2016,creager2019flexibly,rateike2022don,liu2022fair,oh2022learning};
    \item adversarial variational autoencoder~\citep{wu2022semi};
    \item normalizing flows~\citep{balunovic2022fair,cerrato2022fair};
    \item dimensionality reduction~\citep{perez2017fair,samadi2018price,kamani2022efficient,tan2020learning};
    \item residuals~\citep{komiyama2017two};
    \item contrastive learning~\citep{gupta2021controllable};
    \item neural style transfer~\citep{quadrianto2019discovering,quadrianto2018neural}.
\end{itemize}

Another method for improving the fairness of data representations is the removal~\citep{madhavan2020fairness,grgic2018beyond,wang2019approaching} or addition of features~\citep{salazar2021automated,fong2021fairness,Galhotra2022causal}.
\citet{grgic2018beyond} investigated fairness while using different sets of features, thereby making training feature choices.
\citet{madhavan2020fairness} removed discriminating features from the training data.
\citet{salazar2021automated} applied feature creation techniques which apply nonlinear transformation and drop biased features.

\begin{table}[]
\centering
\caption{Publications on In-processing bias mitigation methods.}
\label{table:in}
\begin{adjustbox}{max width=0.45\columnwidth}
\begin{tabular}[t]{llll}
\toprule
Type & Authors [Ref] & Year & Venue \\ \midrule
\multirow{51}{*}{\rotatebox{90}{Regularization}} & \citet{kamiran2010discrimination} & 2010 & ICDM \\
 & \citet{kamishima2011fairness} & 2011 & ICDMW \\
 & \citet{kamishima2012fairness} & 2012 & ECML PKDD \\
 & \citet{ristanoski2013discrimination} & 2013 & CIKM \\
 & \citet{fish2015fair} & 2015 & FATML \\
 & \citet{perez2017fair} & 2017 & ECML PKDD \\
 & \citet{bechavod2017penalizing} & 2017 & arXiv \\
 & \citet{berk2017convex} & 2017 & arXiv \\
 & \citet{quadrianto2017recycling} & 2017 & NeurIPS \\
 & \citet{raff2018fair} & 2018 & AIES \\
 & \citet{enni2018using} & 2018 & ICDM \\
 & \citet{goel2018non} & 2018 & AAAI \\
 & \citet{zhang2019fairness} & 2019 & ICDMW \\
 & \citet{mary2019fairness} & 2019 & ICML \\
 & \citet{beutel2019putting} & 2019 & AIES \\
 & \citet{huang2019stable} & 2019 & ICML \\
 & \citet{aghaei2019learning} & 2019 & AAAI \\
 & \citet{zhang2019faht} & 2019 & IJCAI \\
 & \citet{keya2020equitable} & 2020 & arXiv \\
 & \citet{kim2020fact} & 2020 & ICML \\
 & \citet{jiang2020wasserstein} & 2020 & UAI \\
 & \citet{di2020counterfactual} & 2020 & arXiv \\
 & \citet{abay2020mitigating} & 2020 & arXiv \\
 & \citet{baharlouei2019r} & 2020 & ICLR \\
 & \citet{liu2021fair} & 2020 & Preprint \\
 & \citet{kamani2020multiobjective} & 2020 & Thesis \\
 & \citet{ravichandran2020fairxgboost} & 2020 & arXiv \\
 & \citet{tavakol2020fair} & 2020 & SIGIR \\
 & \citet{romano2020achieving} & 2020 & NeurIPS \\
 & \citet{hickey2020fairness} & 2020 & ECML PKDD \\
 & \citet{wang2021understanding} & 2021 & SIGKDD \\
 & \citet{chuang2021fair} & 2021 & ICLR \\
 & \citet{lowy2021fermi} & 2021 & arXiv \\
 & \citet{zhang2021fair} & 2021 & ICDM \\
 & \citet{grari2021fairnessRenyi} & 2021 & IJCAI \\
 & \citet{yurochkin2021sensei} & 2021 & ICLR \\
 & \citet{zhao2021you} & 2021 & arXiv \\
 & \citet{ranzato2021fairness} & 2021 & CIKM \\
 & \citet{mishler2021fade} & 2021 & arXiv \\
 & \citet{kang2021multifair} & 2021 & arXiv \\
 & \citet{sun2022towards} & 2022 & EuroS\&P \\
 & \citet{zhao2022towards} & 2022 & WSDM \\
 & \citet{wang2022synthesizing} & 2022 & CAV \\
 & \citet{deng2022fifa} & 2022 & arXiv \\
 & \citet{lee2022maximal} & 2022 & Entropy \\
 & \citet{zhang2022longitudinal} & 2022 & AAAI \\
 & \citet{jiang2022generalized} & 2022 & ICLR \\
 & \citet{lee2022maximal2} & 2022 & ICASSP \\
 & \citet{do2022fair} & 2022 & ICML \\
 & \citet{patil2022decorrelation} & 2022 & Future Internet \\
 & \citet{KIM202226} & 2022 & Neurocomputing \\ 
\bottomrule 
\end{tabular}
\end{adjustbox}
\begin{adjustbox}{max width=0.45\columnwidth}
\begin{tabular}[t]{llll}
\toprule
Type & Authors [Ref] & Year & Venue \\ \midrule
\multirow{74}{*}{\rotatebox{90}{Constraints}} & \citet{dwork2012fairness} & 2012 & ITCS \\
 & \citet{calders2013controlling} & 2013 & ICDM \\
 & \citet{fukuchi2015fairness} & 2015 & arXiv \\
 & \citet{fukuchi2015prediction} & 2015 & IEICE Trans. Inf.\& Syst. \\
 & \citet{goh2016satisfying} & 2016 & NeurIPS \\
 & \citet{woodworth2017learning} & 2017 & COLT \\
 & \citet{zafar2017fairness} & 2017 & TheWebConf \\
 & \citet{corbett2017algorithmic} & 2017 & KDD \\
 & \citet{zafar2017fairnesscons} & 2017 & AISTATS \\
 & \citet{komiyama2017two} & 2017 & arXiv \\
 & \citet{zafar2017parity} & 2017 & NeurIPS \\
 & \citet{quadrianto2017recycling} & 2017 & NeurIPS \\
 & \citet{russell2017worlds} & 2017 & NeurIPS \\
 & \citet{kilbertus2017avoiding} & 2017 & NeurIPS \\
 & \citet{agarwal2018reductions} & 2018 & ICML \\
 & \citet{kim2018fairness} & 2018 & NeurIPS \\
 & \citet{narasimhan2018learning} & 2018 & AISTATS \\
 & \citet{gillen2018online} & 2018 & NeurIPS \\
 & \citet{grgic2018beyond} & 2018 & AAAI \\
 & \citet{heidari2018fairness} & 2018 & NeurIPS \\
 & \citet{kearns2018preventing} & 2018 & ICML \\
 & \citet{zhang2018fairness} & 2018 & AAAI \\
 & \citet{gupta2018proxy} & 2018 & arXiv \\
 & \citet{olfat2018spectral} & 2018 & AISTATS \\
 & \citet{zhang2018equality} & 2018 & NeurIPS \\
 & \citet{komiyama2018nonconvex} & 2018 & ICML \\
 & \citet{wu2018fairness} & 2018 & arXiv \\
 & \citet{donini2018empirical} & 2018 & NeurIPS \\
 & \citet{farnadi2018fairness} & 2018 & AIES \\
 & \citet{nabi2018fair} & 2018 & AAAI \\
 & \citet{goel2018non} & 2018 & AAAI \\
 & \citet{wick2019unlocking} & 2019 & NeurIPS \\
 & \citet{celis2019classification} & 2019 & FAccT \\
 & \citet{cotter2019training} & 2019 & ICML \\
 & \citet{balashankar2019fair} & 2019 & arXiv \\
 & \citet{agarwal2019fair} & 2019 & ICML \\
 & \citet{nabi2019learning} & 2019 & ICML \\
 & \citet{cotter2019two} & 2019 & ALT \\
 & \citet{oneto2019taking} & 2019 & AIES \\
 & \citet{cotter2019optimization} & 2019 & JMLR \\
 & \citet{jung2019algorithmic} & 2019 & arXiv \\
 & \citet{lamy2019noise} & 2019 & NeurIPS \\
 & \citet{xu2019achieving} & 2019 & TheWebConf \\
 & \citet{zafar2019fairness} & 2019 & JMLR \\
 & \citet{wang2020robust} & 2020 & NeurIPS \\
 & \citet{chzhen2020minimax} & 2020 & arxiv \\
 & \citet{lohaus2020too} & 2020 & ICML \\
 & \citet{kilbertus2020fair} & 2020 & AISTATS \\
 & \citet{ding2020differentially} & 2020 & AAAI \\
 & \citet{maity2020there} & 2020 & arXiv \\
 & \citet{cho2020fair} & 2020 & NeurIPS \\
 & \citet{padala2020fnnc} & 2020 & IJCAI \\
 & \citet{oneto2020general} & 2020 & IJCNN \\
 & \citet{chzhen2020fair} & 2020 & NeurIPS \\
 & \citet{celis2021fair} & 2021 & PMLR \\
 & \citet{celis2021fairadv} & 2021 & NeurIPS \\
 & \citet{slowik2021algorithmic} & 2021 & arXiv \\
 & \citet{li2021yet} & 2021 & LAK \\
 & \citet{scutari2021achieving} & 2021 & arXiv \\
 & \citet{padh2021addressing} & 2021 & UAI \\
 & \citet{zhang2021omnifair} & 2021 & MOD \\
 & \citet{zhao2021fairness} & 2021 & KDD \\
 & \citet{PETROVIC2021104398} & 2021 & Eng. Appl. Artif. Intell. \\
 & \citet{perrone2021fair} & 2021 & AIES \\
 & \citet{choi2021group} & 2021 & AAAI \\
 & \citet{du2021robust} & 2021 & CIKM \\
 & \citet{lawless2021interpretable} & 2021 & arXiv \\
 & \citet{mishler2021fade} & 2021 & arXiv \\
 & \citet{Park2022privacy} & 2022 & WWW \\
 & \citet{wang2022synthesizing} & 2022 & CAV \\
 & \citet{zhao2022adaptive} & 2022 & KDD \\
 & \citet{boulitsakis2022fairness} & 2022 & arXiv \\
 & \citet{hu2022provably} & 2022 & arXiv \\
 & \citet{wu2022fair} & 2022 & CLeaR \\
 
\bottomrule 
\end{tabular}
\end{adjustbox}
\end{table}

\begin{table}[]
\centering
\caption{Publications on In-processing bias mitigation methods - Part 2.}
\label{table:in2}
\begin{adjustbox}{max width=0.45\columnwidth}
\begin{tabular}[t]{llll}
\toprule
Type & Authors [Ref] & Year & Venue \\ \midrule
\multirow{30}{*}{\rotatebox{90}{Adversarial}} & \citet{beutel2017data} & 2017 & arXiv \\
 & \citet{agarwal2018reductions} & 2018 & ICML \\
 & \citet{gillen2018online} & 2018 & NeurIPS \\
 & \citet{raff2018gradient} & 2018 & DSAA \\
 & \citet{wadsworth2018achieving} & 2018 & arXiv \\
 & \citet{kearns2018preventing} & 2018 & ICML \\
 & \citet{zhang2018mitigating} & 2018 & AIES \\
 & \citet{adel2019one} & 2019 & AAAI \\
 & \citet{beutel2019putting} & 2019 & AIES \\
 & \citet{sadeghi2019global} & 2019 & ICCV \\
 & \citet{zhao2019inherent} & 2019 & NeurIPS \\
 & \citet{xu2019fairgan} & 2019 & Big Data \\
 & \citet{grari2019fair} & 2019 & ICDM \\
 & \citet{celis2019improved} & 2019 & arXiv \\
 & \citet{garcia2020reducing} & 2020 & SMU DSR \\
 & \citet{Yurochkin2020Training} & 2020 & ICLR \\
 & \citet{roh2020fr} & 2020 & ICML \\
 & \citet{delobelle2020ethical} & 2020 & ASE \\
 & \citet{rezaei2020fairness} & 2020 & AAAI \\
 & \citet{lahoti2020fairness} & 2020 & NeurIPS \\
 & \citet{grari2021fairness} & 2021 & arXiv \\
 & \citet{grari2021fairnessRenyi} & 2021 & IJCAI \\
 & \citet{amend2021improving} & 2021 & JCSC \\
 & \citet{rezaei2021robust} & 2021 & AAAI \\
 & \citet{chen2022fair} & 2022 & arXiv \\
 & \citet{liang2022joint} & 2022 & arXiv \\
 & \citet{guanhongfse22} & 2022 & FSE \\
 & \citet{petrovic2022fair} & 2022 & Neurocomputing \\
 & \citet{yang2022algorithmic} & 2022 & medRxiv \\
 & \citet{yazdani2022distraction} & 2022 & arXiv \\ \midrule
\multirow{17}{*}{\rotatebox{90}{Compositional}} & \citet{calders2010three} & 2010 & DMKD \\
 & \citet{pleiss2017fairness} & 2017 & NeurIPS \\
 & \citet{dwork2018decoupled} & 2018 & FAccT \\
 & \citet{ustun2019fairness} & 2019 & ICML \\
 & \citet{oneto2019taking} & 2019 & AIES \\
 & \citet{iosifidis2019fae} & 2019 & Big Data \\
 & \citet{monteirofair2021} & 2021 & PLM \\
 & \citet{ranzato2021fairness} & 2021 & CIKM \\
 & \citet{mishler2021fade} & 2021 & arXiv \\
 & \citet{kobayashi2021one} & 2021 & DiTTEt \\
 & \citet{jin2022input} & 2022 & ICML \\
 & \citet{zhenpengmaat2022} & 2022 & FSE \\
 & \citet{roy2022multi} & 2022 & DS \\
 & \citet{liu2022accuracy} & 2022 & CMS \\
 & \citet{blanzeisky2022using} & 2022 & Knowl Eng Rev \\
 & \citet{boulitsakis2022fairness} & 2022 & arXiv \\
 & \citet{suriyakumar2022personalization} & 2022 & arXiv \\
\bottomrule 
\end{tabular}
\end{adjustbox}
\begin{adjustbox}{max width=0.47\columnwidth}
\begin{tabular}[t]{llll}
\toprule
Type & Authors [Ref] & Year & Venue \\ \midrule
\multirow{60}{*}{\rotatebox{90}{Adjusted}} & \citet{luo2015discrimination} & 2015 & DaWaK \\
 & \citet{joseph2016fairness} & 2016 & NeurIPS \\
 & \citet{johnson2016impartial} & 2016 & Stat Sci \\
 & \citet{kusner2017counterfactual} & 2017 & NeurIPS \\
 & \citet{joseph2018meritocratic} & 2018 & AIES \\
 & \citet{hashimoto2018fairness} & 2018 & ICML \\
 & \citet{madras2018predict} & 2018 & NeurIPS \\
 & \citet{alabi2018unleashing} & 2018 & COLT \\
 & \citet{hebert2018multicalibration} & 2018 & ICML \\
 & \citet{chiappa2018causal} & 2018 & IFIP \\
 & \citet{kilbertus2018blind} & 2018 & ICML \\
 & \citet{kamishima2018model} & 2018 & DMKD \\
 & \citet{dimitrakakis2019bayesian} & 2019 & AAAI \\
 & \citet{chiappa2019path} & 2019 & AAAI \\
 & \citet{noriega2019active} & 2019 & AIES \\
 & \citet{chakraborty2019software} & 2019 & arXiv \\
 & \citet{madras2019fairness} & 2019 & FAccT \\
 & \citet{iosifidis2019adafair} & 2019 & CIKM \\
 & \citet{mandal2020ensuring} & 2020 & NeurIPS \\
 & \citet{kilbertus2020fair} & 2020 & AISTATS \\
 & \citet{martinez2020minimax} & 2020 & ICML \\
 & \citet{iosifidis2020mathsf} & 2020 & DS \\
 & \citet{liu2021fair} & 2020 & Preprint \\
 & \citet{hu2020fairnn} & 2020 & DS \\
 & \citet{da2020fairness} & 2020 & Thesis \\
 & \citet{chakraborty2020fairway} & 2020 & FSE \\
 & \citet{kamani2020multiobjective} & 2020 & Thesis \\
 & \citet{zhang2020learning} & 2020 & arXiv \\
 & \citet{ignatiev2020towards} & 2020 & CP \\
 & \citet{fairn2021sharma} & 2021 & AIES \\
 & \citet{ezzeldin2021fairfed} & 2021 & arXiv \\
 & \citet{wang2021fair} & 2021 & FAccT \\
 & \citet{ozdayi2021bifair} & 2021 & arXiv \\
 & \citet{zhang2021farf} & 2021 & PAKDD \\
 & \citet{perrone2021fair} & 2021 & AIES \\
 & \citet{islam2021can} & 2021 & AIES \\
 & \citet{roh2021fairbatch} & 2021 & ICLR \\
 & \citet{hort2021did} & 2021 & ASE \\
 & \citet{valdivia2021fair} & 2021 & Int. J. Intell. Syst. \\
 & \citet{lee2021fair} & 2021 & ICML \\
 & \citet{cruz2021bandit} & 2021 & ICDM \\
 & \citet{roy2022learning} & 2022 & ECML PKDD \\
 & \citet{wang2022mitigating} & 2022 & arXiv \\
 & \citet{sikdar2022getfair} & 2022 & FAccT \\
 & \citet{Agarwal2022power} & 2022 & FAccT \\
 & \citet{Park2022privacy} & 2022 & WWW \\
 & \citet{djebrouni2022towards} & 2022 & Eurosys \\
 & \citet{Iosifidis2022parity} & 2022 & KAIS \\
 & \citet{SHORT2022} & 2022 & 	Int. J. Forecast. \\
 & \citet{maheshwari2022fairgrad} & 2022 & arXiv \\
 & \citet{zhao2022adaptive} & 2022 & KDD \\
 & \citet{tizpaz2022fairness} & 2022 & ICSE \\
 & \citet{roy2022multi} & 2022 & DS \\
 & \citet{mohammadi2022feta} & 2022 & arXiv \\
 & \citet{gao2022fair} & 2022 & ICSE \\
 & \citet{huang2022fair} & 2022 & Expert Syst. Appl. \\
 & \citet{candelieri2022fair} & 2022 & arXiv \\
 & \citet{anahideh2022fair} & 2022 & Expert Syst. Appl. \\
 & \citet{rateike2022don} & 2022 & FAccT \\
 & \citet{li2022accurate} & 2022 & arXiv \\ 
\bottomrule 
\end{tabular}
\end{adjustbox}
\end{table}

\subsection{In-processing Bias Mitigation Methods}
\label{section:in}
This section presents in-processing methods; methods that mitigate bias during the training procedure of the algorithm.
Overall, we found a total of \pubin publications (see Table~\ref{table:in}, Table~\ref{table:in2} for more details) that apply in-processing methods.
For more details on in-processing methods, we refer to the survey by \citet{surveyIN}, which provides information on 38 in-processing approaches developed for various ML tasks.

\subsubsection{Regularization and Constraints}
\label{section:in-loss}
Regularization and constraints are both approaches that apply changes to the learning algorithm's loss function.
Regularization adds a term to the loss function.
While the original loss function is based on accuracy metrics, the purpose of a regularization term is to penalize discrimination (i.e., discrimination leads to a higher loss of the ML algorithm). 
Constraints on the other hand determine specific bias levels (according to loss functions) that cannot be breached during training. 

To widen the range of fairness definitions that can be considered when applying constraints, \citet{celis2019classification} proposed a Meta-algorithm.
This Meta-algorithm takes a fairness constraint as input.

When applied to Decision Trees, regularization can be used to modify the splitting criteria~\citep{kamiran2010discrimination,zhang2019faht,ranzato2021fairness,zhang2019fairness,zhang2021fair,zhang2022longitudinal,wang2022synthesizing}. 
Traditionally, leaves are iteratively split to achieve an improvement in accuracy.
To improve fairness while training, \citet{kamiran2010discrimination} considered fairness in addition to accuracy when leaf splitting.
They applied three splitting strategies: 
\begin{enumerate}
    \item only allow non-discriminatory splits;
    \item choose best split according to $\delta_{accuracy}/ \delta_{discrimination}$;
    \item choose best split according to $\delta_{accuracy} + \delta_{discrimination}$.
\end{enumerate}


While constraints and regularization usually utilize group fairness definitions, they have also been applied for achieving individual fairness~\citep{gillen2018online,kim2018fairness,dwork2012fairness,jung2019algorithmic}.
Moreover, they can be applied to achieve fairness for multiple sensitive attributes and fairness definitions~\citep{kang2021multifair,tavakol2020fair,kang2021multifair,padh2021addressing,komiyama2018nonconvex,kearns2018preventing}, or extend existing adjustments, such as adding fairness regularization in addition to the L2 norm, which is used to avoid overfitting~\citep{kamishima2012fairness,kamishima2011fairness}.

\subsubsection{Adversarial Learning}
\label{section:in-adversarial}
Adversarial learning simultaneously trains classification models and their adversaries~\citep{dalvi2004adversarial}.
While the classification model is trained to predict ground truth values, the adversary is trained to exploit fairness issues.
Both models then compete against each other, to improve their performance.


\citet{zhang2018mitigating} trained a Logistic Regression model to predict the label $Y$ while preventing an adversary from predicting the protected attribute under consideration of three fairness metrics: Demographic Parity, Equality of Odds, and Equality of Opportunity.
Both, predictor and adversary, are implemented as Logistic regression models.

Similarly, \citet{beutel2017data} trained a neural network to predict two outputs: labels and sensitive attributes.
While a high overall accuracy is desired, the adversarial setting reduces the ability to predict sensitive information.
The network is designed to share layers between the two output, such that only one model is trained~\citep{adel2019one,sadeghi2019global,delobelle2020ethical,beutel2019putting,raff2018gradient}.

\citet{lahoti2020fairness} proposed Adversarially Reweighted Learning (ARL) in which a learner is trained to optimize performance on a classification task while the adversary adjusts the weights of computationally-identifiable regions in the input space with high training loss.
By so-doing, the learner can then improve performance in these regions.

Other than using adversaries to prevent the ability to predict sensitive attributes (e.g., for reducing bias according to population groups), it has also been used to improve robustness to data poisoning~\citep{roh2020fr}, to improve individual fairness~\citep{Yurochkin2020Training}, and to reweigh training data~\citep{petrovic2022fair}.
In particular, \citet{petrovic2022fair} used adversarial training to learn a reweighing function for training data instances as an in-processing procedure (contrary to applying reweighing as pre-processing, see Section \ref{section:pre-sampling}).

\subsubsection{Compositional}
\label{section:in-comp}
Compositional approaches combat bias by training multiple classification models.
Predictions can then be made by a specific classification model for each population group (e.g., privileged and unprivileged)~\citep{calders2010three,pleiss2017fairness,ustun2019fairness,oneto2019taking,boulitsakis2022fairness,jin2022input,suriyakumar2022personalization} or in an ensemble fashion (i.e., a voting of multiple classification models at the same time)~\citep{mishler2021fade,roy2022multi,liu2022accuracy,kobayashi2021one,monteiroproposal,ranzato2021fairness,iosifidis2019fae,zhenpengmaat2022}.

While decoupled classification models for privileged and unprivileged groups can achieve improved accuracy for each group, the amount training data for each classifier is reduced. 
To reduce the impact of small training data sizes \citet{dwork2018decoupled} utilised transfer training. With their transfer learning approach, they trained classifiers on data for the respective group and data from the other groups with reduced weight. 
\citet{ustun2019fairness} built upon the work of \citet{dwork2018decoupled} and incorporated ``preference guarantees'', which states that each group prefers their decoupled classifier over a classifier trained on all training data and any classifier of the other groups. Similarly, \citet{suriyakumar2022personalization} followed the concept of ``fair use'', which states that if a classification uses sensitive group information, it should improve performance for every group.

Training multiple classification models with different fairness goals allows for the creation of a pareto-front of solutions~\citep{roy2022multi,liu2022accuracy,mishler2021fade,valdivia2021fair,blanzeisky2022using}. 
Practitioners can then choose which fairness-accuracy trade-off best suits their need.
For example, \citet{liu2022accuracy} treated bias mitigation as multi-objective optimization problem that explores fairness-accuracy trade-offs under consideration of multiple fairness metrics.
\citet{mishler2021fade} proposed an ensemble method that builds classification models based on a weighted combination of metrics chosen by users.

\subsubsection{Adjusted Learning}
\label{section:in-adjusted}
Adjusted learning methods mitigate bias via changing the learning procedure of algorithms or the creation of novel algorithms~\citep{dunkelau2019fairness}. 
Changes have been suggested for a variety of classification models, including Bayesian models~\citep{kamishima2013independence,dimitrakakis2019bayesian}, Markov Random Fields~\citep{zhang2020learning}, Neural Networks~\citep{martinez2020minimax,raff2018gradient,hu2020fairnn}, Decision Trees, bandits~\citep{joseph2018meritocratic,joseph2016fairness,auer2002finite}, boosting~\citep{roy2022multi,iosifidis2019adafair,iosifidis2020mathsf,hebert2018multicalibration}, Logistic Regression~\citep{roh2021fairbatch}.
We outline a selection of publications in the following, to provide insight on techniques applied to different classification models.

\citet{noriega2019active} proposed an active learning framework for training Decision Trees. During training, a decision maker is able to collect more information about individuals to achieve fairness in predictions.
In this context, not all information about individuals is available. There is an information budget that determines how many enquiries can be performed.
Similarly, \citet{anahideh2022fair} used an active learning framework to balance accuracy and fairness by selecting instances to be labelled.

\citet{madras2018predict} proposed a rejection learning approach for joint decision-making with classification models and external decision makers.
In particular, the classification model learns when to defer from making prediction (i.e., when it is more useful to have predictions from external decision makers).
If the coverage of classification can be reduced (i.e., the classification model abstains from making some of the predictions), selective classification approaches can be used~\citep{lee2021fair}.

\citet{martinez2020minimax} proposed the algorithm Approximate Projection onto Star Sets (APStar) to train Deep Neural Networks to minimize the maximum risk among all population groups.
This procedure ensures that the final classifier is part of the Pareto Front~\citep{emmerich2018tutorial}.
\citet{hu2020fairnn} incorporated representation learning into the training procedure of Neural Networks to learn them jointly the classifier.

\citet{hebert2018multicalibration} proposed \textit{Multicalibration}, a learning procedure similar to boosting.
A classifier is trained iteratively. At each iteration, the predictions of the most biased subgroup are corrected until the classifier is adequately calibrated.

\citet{hashimoto2018fairness} found fairness issues with the use of empirical risk minimization and proposed the use of distributionally robust optimization (DRO) when training classifiers such as Logistic Regression. During training, DRO optimizes the worst-case risk over all groups present.

\citet{kilbertus2018blind} adjusted the training procedure for Logistic Regression to take privacy into account. Sensitive user information is encrypted such that it cannot be used for classification tasks while retaining the ability to verify fairness issues.
By doing so, users can provide sensitive information without the fear that someone can read them.

The learning procedure of existing classification models has also been adjusted by tuning their hyper-parameters~\citep{chakraborty2020fairway,perrone2021fair,valdivia2021fair,cruz2021bandit,hort2021did,da2020fairness,tizpaz2022fairness,islam2021can,chakraborty2019software}.

\subsection{Post-processing Bias Mitigation Methods}
\label{section:post}
Post-processing bias mitigation methods are applied once a classification model has been successfully trained.
With \pubpost publications that apply post-processing methods (Table~\ref{table:post}), post-processing methods are the least frequently applied of those covered in this survey.

\begin{table}[]
\centering
\caption{Publications on Post-processing bias mitigation methods.}
\label{table:post}
\begin{adjustbox}{max width=0.45\columnwidth}
\begin{tabular}[t]{llll}
\toprule
Type & Authors [Ref] & Year & Venue  \\ \midrule

\multirow{2}{*}{Input}  & \citet{adler2018auditing} & 2018 & KAIS \\
 & \citet{li2022training} & 2022 & ICSE \\ \midrule
\multirow{26}{*}{\rotatebox{90}{Output}} & \citet{pedreschi2009measuring} & 2009 & SDM \\
 & \citet{kamiran2012decision} & 2012 & ICDM \\
 & \citet{fish2015fair} & 2015 & FATML \\
 & \citet{fish2016confidence} & 2016 & SDM \\
 & \citet{liu2018fairmod} & 2018 & arXiv \\
 & \citet{kim2018fairness} & 2018 & NeurIPS \\
 & \citet{zhang2018achieving} & 2018 & IJCAI \\
 & \citet{kamiran2018exploiting} & 2018 & 	J. Inf. Sci. \\
 & \citet{menon2018cost} & 2018 & FAccT \\
 & \citet{chzhen2019leveraging} & 2019 & NeurIPS \\
 & \citet{chiappa2019path} & 2019 & AAAI \\
 & \citet{iosifidis2019fae} & 2019 & Big Data \\
 & \citet{lohia2019post} & 2019 & ICASSP \\
 & \citet{wei2020optimized} & 2020 & PMLR \\
 & \citet{alabdulmohsin2020fair} & 2020 & arXiv \\
 & \citet{alabdulmohsin2021near} & 2021 & NeurIPS \\
 & \citet{lohia2021priority} & 2021 & arXiv \\
 & \citet{NGUYEN2021542} & 2021 & 	J. Inf. Sci. \\
 & \citet{kobayashi2021one} & 2021 & DiTTEt \\
 & \citet{jang2022group} & 2022 & AAAI \\
 & \citet{pentyala2022privfairfl} & 2022 & arXiv \\
 & \citet{snel2022practical} & 2022 & Com. Soc. Res. J. \\
 & \citet{alghamdi2022beyond} & 2022 & arXiv \\
 & \citet{mohammadi2022feta} & 2022 & arXiv \\
 & \citet{zeng2022fair2} & 2022 & arXiv \\
 & \citet{zeng2022bayes} & 2022 & arXiv \\

\bottomrule 
\end{tabular}
\end{adjustbox}
\begin{adjustbox}{max width=0.45\columnwidth}
\begin{tabular}[t]{llll}
\toprule
Type & Authors [Ref] & Year & Venue \\ \midrule
\multirow{28}{*}{\rotatebox{90}{Classifier}} & \citet{calders2010three} & 2010 & DMKD \\
 & \citet{kamiran2010discrimination} & 2010 & ICDM \\
 & \citet{hardt2016equality} & 2016 & NeurIPS \\
 & \citet{woodworth2017learning} & 2017 & COLT \\
 & \citet{pleiss2017fairness} & 2017 & NeurIPS \\
 & \citet{gupta2018proxy} & 2018 & arXiv \\
 & \citet{morina2019auditing} & 2019 & arXiv \\
 & \citet{noriega2019active} & 2019 & AIES \\
 & \citet{edit2019fairness} & 2019 & JSAI \\
 & \citet{kim2019multiaccuracy} & 2019 & AIES \\
 & \citet{chzhen2020fairPlug} & 2020 & NeurIPS \\
 & \citet{chzhen2020minimax} & 2020 & arxiv \\
 & \citet{savani2020intra} & 2020 & NeurIPS \\
 & \citet{awasthi2020equalized} & 2020 & PMLR \\
 & \citet{kim2020fact} & 2020 & ICML \\
 & \citet{jiang2020wasserstein} & 2020 & UAI \\
 & \citet{chzhen2020fair} & 2020 & NeurIPS \\
 & \citet{du2021fairness} & 2021 & NeurIPS \\
 & \citet{schreuder2021classification} & 2021 & UAI \\
 & \citet{mishler2021fairness} & 2021 & FAccT \\
 & \citet{mishler2021fade} & 2021 & arXiv \\
 & \citet{kanamori2021fairness} & 2021 & JSAI \\
 & \citet{grabowicz2022marrying} & 2022 & FAccT \\
 & \citet{Iosifidis2022parity} & 2022 & KAIS \\
 & \citet{mehrabi2022attributing} & 2022 & TrustNLP \\
 & \citet{zhang2022fair} & 2022 & FairWARE \\
 & \citet{wu2022fairness} & 2022 & FAccT \\
 & \citet{marcinkevics2022debiasing} & 2022 & MLHC \\
\bottomrule 
\end{tabular}
\end{adjustbox}
\end{table}

\subsubsection{Input Correction}
\label{section:post-in}
Input correction approaches apply a modification step to the testing data. This is comparable to pre-processing approaches (Section \ref{section:pre})~\citep{dunkelau2019fairness}, which conduct modifications to training data (e.g., relabelling, perturbation and representation learning).

We found only two publications that applied input corrections to testing data, both of which used perturbations.
While \citet{adler2018auditing} used perturbation in a post-processing stage, \citet{li2022training} first performed perturbation in a pre-processing stage and then applied an identical procedure for post-processing.

\subsubsection{Classifier Correction}
\label{section:post-clf}
Post-processing approaches can also directly be applied to classification models, which \citet{savani2020intra} called intra-processing. A successfully trained classification model is adapted to obtain a fairer one.
Such modification have been applied to Naive Bayes~\citep{calders2010three}, Logistic Regression~\citep{jiang2020wasserstein}, Decision Trees~\citep{kamiran2010discrimination,kanamori2021fairness,zhang2022fair}, Neural Networks~\citep{du2021fairness,mehrabi2022attributing,savani2020intra,marcinkevics2022debiasing} and Regression Models~\citep{chzhen2020fair}.

\citet{hardt2016equality} proposed the modification of classifiers to achieve fairness with respect to Equalized Odds and Equality of Opportunity.
Given an unfair classifier $\widehat{Y}$, the classifier $\widetilde{Y}$ is derived by solving an optimization problem under consideration of fairness loss terms.
This approach has been adapted and extended by further publications~\citep{gupta2018proxy,morina2019auditing,awasthi2020equalized,mishler2021fairness}.

\citet{woodworth2017learning} showed that this kind of modification can lead to a poor accuracy, for example when the loss function is not strictly convex. In addition to constraints during training, they proposed an adaptation of the approach by \citet{hardt2016equality}.

\citet{pleiss2017fairness} split a classifier in two ($h_0$, $h_1$), for the privileged and unprivileged group.
To balance the false positive and false negative rate of the two classifiers, $h_1$ is adjusted such that with a probability of $\alpha$ the class mean is returned rather than the actual prediction. 
\citet{noriega2019active} followed the calibration approach of \citet{pleiss2017fairness}.

\citet{kamiran2010discrimination} modified Decision Tree classifiers by relabeling leaf nodes.
The goal of relabeling was to reduce bias while sacrificing as little accuracy as possible.
A greedy procedure was followed which iteratively selects the best leaf to relabel (i.e., highest ratio of fairness improvement per accuracy loss).
\citet{kanamori2021fairness} formulated the modification of branching thresholds for Decision Trees as a mixed integer program.

\citet{kim2019multiaccuracy} proposed \textit{Multiaccuracy Boost}, a post-processing approach similar to boosting for training classifiers.
Given a black-box classifier and a learning algorithm, \textit{Multiaccuracy Boost} iteratively adapts the current classifier based on its predictive performance.



\subsubsection{Output Correction}
\label{section:post-out}
The latest stage of applying bias mitigation methods is the correction of the output. In particular, the predicted labels are modified.

\citet{pedreschi2009measuring} considered the correction of rule-based classifiers, such as CPAR~\citep{yin2003cpar}. 
For each individual, the $k$ rules with highest confidence are selected to determine the probability for each output label.
Given that some of the rules can be discriminatory, their confidence level is adjusted to reduce biased labels.

\citet{menon2018cost} proposed a plugin approach for thresholding predictions. 
To determine the thresholds to use, the class probabilities are estimated using logistic regression.

\citet{kamiran2012decision,kamiran2018exploiting} introduced the notion of reject option which modifies the prediction of individuals close to the decision boundary.
In particular, individuals belonging to the unprivileged group receive a positive outcome and privileged individuals an unfavourable outcome. 
Similarly, \citet{lohia2019post} relabeled individuals that are likely to receive biased outcomes, but rather than considering the decision boundary, they used an ``individual bias detector'' to find predictions that are likely suffer from individual discrimination. 
This work was extended in 2021, where individuals were ranked based on their ``Unfairness Quotient'' (i.e., the difference between regular prediction and with perturbed protected attribute). 
\citet{fish2016confidence} proposed a confidence-based approach which returns a positive label for each prediction above a given threshold. 
This has also been applied to AdaBoost~\citep{fish2015fair}.
Other than using a general threshold for all instances, group dependent thresholds can be used~\citep{chzhen2019leveraging,iosifidis2019fae,kobayashi2021one,zeng2022bayes,zeng2022fair2,jang2022group,pentyala2022privfairfl,alabdulmohsin2020fair}. 

\citet{chiappa2019path} addressed the fairness of causal models under consideration of a counterfactual world in which individuals belong to a different population group.
The impact of the protected attribute on the prediction outcome is corrected to ensure that it coincides with counterfactual predictions.
This way, sensitive information is removed while other information remains unchanged.

\subsection{Combined Approaches}
\label{section:combined}
While most publications proposed the use of a single type of bias mitigation method, we found \pubmulti that applied multiple techniques at the same time (e.g., two pre-processing methods, one in-processing and one post-processing methods).
Table~\ref{table:combined} summarizes these approaches.

Among these \pubmulti publications, 86\% (60 out of \pubmulti) applied in-processing, 54\% (38 out of \pubmulti) applied pre-processing, and 31\% (22 out of \pubmulti) applied post-processing methods.

Additionally, 26 out of \pubmulti publications applied multiple types of bias mitigation methods at the same stage of the development process (e.g., two pre-processing approaches).
In particular, the are 7 publications which applied multiple pre-processing methods.
Among these 7 publications, 5 applied sampling and relabeling~\citep{vzliobaite2011handling,kamiran2012data,calders2009building,iosifidis2019fairness,sun2022towards}.
The remaining 19 out of 26 publications applied multiple in-processing methods, 17 of which include regularization or constraints.

47 publications applied at least two methods at different stages of the development process for ML models (e.g., one pre-processing and one in-processing method). 
This illustrates that bias mitigation methods can be used in conjunction~\citep{ghai2022cascaded}.
Moreover, there are three publications that addressed bias mitigation at each stage: pre-processing, in-processing and post-processing~\citep{calders2010three,gupta2018proxy,iosifidis2019fae}.

\citet{calders2010three} proposed three approaches for achieving discrimination-free classification of naive bayes models.
At first, a latent variable is added to represent unbiased labels. The data is then used to train a model for each possible sensitive attribute value.
Lastly, the probabilities output by the model are modified to account for unfavourable treatment (i.e., increasing the probability of positive outcomes for the unprivileged group and reducing it for the privileged group).

\citet{gupta2018proxy} tackled the problem of bias mitigation for situations where group labels are missing in the datasets. To combat this issue, they created a latent ``proxy'' variable for the group membership and incorporated constraints for achieving fairness for such proxy groups in the training procedure.
Lastly, they followed the approach of \citet{hardt2016equality} to debias and existing classifier by adding an additional variable to the prediction problem (see Section \ref{section:post-clf}).

\citet{iosifidis2019fae} followed an ensemble approach of multiple AdaBoost classifiers. In particular, each classifier is trained on an equal amount of instances from each population group and label by sampling.
Predictions are then modified by applying group-dependent thresholds.

\begin{table}[]
\centering
\caption{Publications with multiple bias mitigation methods. ``X'' indicates that the publication applies a bias mitigation approach of the corresponding category.}
\label{table:combined}
\begin{adjustbox}{max width=0.48\columnwidth}
\begin{tabular}[t]{lccc}
\toprule
                  & \multicolumn{3}{l}{Processing Method} \\
Authors  & Pre        & In         & Post        \\ \midrule
\citet{sun2022towards} & x x & x & \\
 \citet{calders2009building} & x x &  & \\
 \citet{vzliobaite2011handling} & x x &  & \\
 \citet{hajian2012methodology} & x x &  & \\
 \citet{kamiran2012data} & x x &  & \\
 \citet{iosifidis2019fairness} & x x &  & \\
 \citet{chakraborty2022fair} & x x &  & \\
 \citet{oneto2019taking} & x & x x & \\
 \citet{calders2010three} & x & x & x\\
 \citet{gupta2018proxy} & x & x & x\\
 \citet{iosifidis2019fae} & x & x & x\\
 \citet{perez2017fair} & x & x & \\
 \citet{komiyama2017two} & x & x & \\
 \citet{kilbertus2017avoiding} & x & x & \\
 \citet{grgic2018beyond} & x & x & \\
 \citet{madras2019fairness} & x & x & \\
 \citet{xu2019fairgan} & x & x & \\
 \citet{abay2020mitigating} & x & x & \\
 \citet{hu2020fairnn} & x & x & \\
 \citet{chakraborty2020fairway} & x & x & \\
 \citet{chuang2021fair} & x & x & \\
 \citet{zhang2021farf} & x & x & \\
 \citet{grari2021fairness} & x & x & \\
 \citet{du2021robust} & x & x & \\
 \citet{amend2021improving} & x & x & \\
 \citet{cruz2021bandit} & x & x & \\
 \citet{chen2022fair} & x & x & \\
 \citet{liang2022joint} & x & x & \\
 \citet{Agarwal2022power} & x & x & \\
 \citet{zhenpengmaat2022} & x & x & \\
 \citet{wu2022fair} & x & x & \\
 \citet{rateike2022don} & x & x & \\
 \citet{KIM202226} & x & x & \\
 \citet{suriyakumar2022personalization} & x & x & \\
 \citet{zhang2018achieving} & x &  & x\\
 \citet{wei2020optimized} & x &  & x\\
 \citet{pentyala2022privfairfl} & x &  & x\\
 \citet{li2022training} & x &  & x\\
\bottomrule 
\end{tabular}
\end{adjustbox}
\hfill
\begin{adjustbox}{max width=0.48\columnwidth}
\begin{tabular}[t]{lccc}
\toprule
                  & \multicolumn{3}{l}{Processing Method} \\
Authors  & Pre        & In         & Post        \\ \midrule
 \citet{mishler2021fade} &  & x x x & x\\
 \citet{quadrianto2017recycling} &  & x x & \\
 \citet{agarwal2018reductions} &  & x x & \\
 \citet{gillen2018online} &  & x x & \\
 \citet{kearns2018preventing} &  & x x & \\
 \citet{goel2018non} &  & x x & \\
 \citet{beutel2019putting} &  & x x & \\
 \citet{kilbertus2020fair} &  & x x & \\
 \citet{liu2021fair} &  & x x & \\
 \citet{kamani2020multiobjective} &  & x x & \\
 \citet{perrone2021fair} &  & x x & \\
 \citet{grari2021fairnessRenyi} &  & x x & \\
 \citet{ranzato2021fairness} &  & x x & \\
 \citet{Park2022privacy} &  & x x & \\
 \citet{wang2022synthesizing} &  & x x & \\
 \citet{zhao2022adaptive} &  & x x & \\
 \citet{roy2022multi} &  & x x & \\
 \citet{boulitsakis2022fairness} &  & x x & \\
 \citet{kamiran2010discrimination} &  & x & x\\
 \citet{fish2015fair} &  & x & x\\
 \citet{woodworth2017learning} &  & x & x\\
 \citet{pleiss2017fairness} &  & x & x\\
 \citet{kim2018fairness} &  & x & x\\
 \citet{chiappa2019path} &  & x & x\\
 \citet{noriega2019active} &  & x & x\\
 \citet{chzhen2020minimax} &  & x & x\\
 \citet{kim2020fact} &  & x & x\\
 \citet{jiang2020wasserstein} &  & x & x\\
 \citet{chzhen2020fair} &  & x & x\\
 \citet{kobayashi2021one} &  & x & x\\
 \citet{Iosifidis2022parity} &  & x & x\\
 \citet{mohammadi2022feta} &  & x & x\\

\bottomrule 
\end{tabular}
\end{adjustbox}
\end{table}

\subsection{Classification Models}
Here we outline the classification models on which the three types of bias mitigation methods (pre-, in-, post-processing) have been applied on.
Table \ref{table:clfs} shows the frequency with which each type of classification model has been applied.

Currently, the most frequently used classification model is Logistic Regression, for each method type (pre-, in-, post-processing), with a total of 140 unique publications using it for their experiments.
The second most frequently used classification models are Neural Networks (NNs). 
A total of 102 publication used NNs for their experiments, with the majority being in-processing methods.
Linear Regression models have been used in 22 publications.

Decision Trees (36 publications) and Random Forests (45 publications) are also frequently used. Moreover, different Decision Tree variants have been used, such as Hoeffding trees, C4.5, J48 and Bayesian random forests.

While the range of classification models is diverse, some of them are similar to one another:
\begin{itemize}
    \item Boosting: AdaBoost, XGBoost, SMOTEBoost, Boosting, LightGBM, OSBoost, Gradient Tree Boosting, CatBoost;
    \item Rule-based: RIPPER, PART, CBA, Decision Set, Rule Sets, Decision Rules.
\end{itemize}

\begin{table}[]
\centering
\caption{Frequency of classification model usage for evaluating bias mitigation methods. Amounts are provided for each category and as a unique measure to avoid counting publications with multiple approaches double.}
\label{table:clfs}
\begin{adjustbox}{max width=0.45\columnwidth}
\begin{tabular}[t]{lrrrr}
\toprule
                 &  & \multicolumn{3}{c}{Method} \\
Model & Unique & Pre        & In         & Post        \\ \midrule
 Logistic Regression  & 140 & 58 & 80 & 19 \\
 Neural Network  & 102 & 34 & 65 & 17 \\
 Random Forest  & 45 & 20 & 22 & 14 \\
 SVM  & 37 & 15 & 18 & 9 \\
 Decision Tree  & 36 & 14 & 16 & 9 \\
 Naive Bayes  & 24 & 12 & 11 & 5 \\
 Linear Regression  & 22 & 4 & 20 & 3 \\
 Nearest Neighbor  & 13 & 7 & 2 & 5 \\
 AdaBoost  & 8 & 1 & 5 & 4 \\
 XGBoost  & 8 & 1 & 6 & 1 \\
 Causal  & 7 & 2 & 6 & 1 \\
 LightGBM  & 4 & 2 & 3 & 0 \\
 Bandit  & 3 & 0 & 3 & 0 \\
 Boosting  & 3 & 0 & 2 & 2 \\
 J48  & 2 & 1 & 1 & 0 \\
 Bayesian  & 2 & 0 & 1 & 1 \\
 Hoeffding Tree  & 2 & 1 & 1 & 0 \\
 Gaussian Process  & 2 & 2 & 0 & 0 \\
 CPAR  & 1 & 0 & 0 & 1 \\
 RIPPER  & 1 & 1 & 0 & 0 \\
 PART  & 1 & 1 & 0 & 0 \\
 C4.5  & 1 & 1 & 0 & 0 \\
CBA  & 1 & 0 & 1 & 0 \\
\bottomrule 
\end{tabular}
\hfill
\end{adjustbox}
\begin{adjustbox}{max width=0.45\columnwidth}
\begin{tabular}[t]{lrrrr}
\toprule
                 &  & \multicolumn{3}{c}{Method} \\
Model & Unique & Pre        & In         & Post        \\ \midrule
 Lattice  & 1 & 1 & 1 & 1 \\
 Lasso  & 1 & 0 & 1 & 0 \\
 PSL  & 1 & 0 & 1 & 0 \\
 BART  & 1 & 0 & 1 & 0 \\
 RTL  & 1 & 0 & 1 & 0 \\
 Tree Ensemble  & 1 & 0 & 1 & 0 \\
 AUE  & 1 & 1 & 0 & 0 \\
 CART  & 1 & 0 & 1 & 0 \\
 SMOTEBoost  & 1 & 0 & 1 & 0 \\
 Gradient boosted trees  & 1 & 1 & 0 & 1 \\
 Cox model  & 1 & 0 & 1 & 0 \\
 Decision Rules  & 1 & 0 & 1 & 0 \\
 Gradient Tree Boosting  & 1 & 0 & 1 & 0 \\
 Kmeans  & 1 & 0 & 1 & 0 \\
 OSBoost  & 1 & 0 & 1 & 0 \\
 POEM  & 1 & 0 & 1 & 0 \\
 Markov random filed  & 1 & 0 & 1 & 0 \\
 SMSGDA  & 1 & 0 & 1 & 0 \\
 Probabilistic circuits  & 1 & 0 & 1 & 0 \\
 Rule Sets  & 1 & 0 & 1 & 0 \\
 Ridge Regression  & 1 & 0 & 1 & 1 \\
 Extreme Random Forest  & 1 & 1 & 0 & 0 \\
 Factorization Machine  & 1 & 1 & 0 & 0 \\
 Discriminant analysis  & 1 & 0 & 1 & 0 \\
 Generalized Linear Model  & 1 & 0 & 1 & 0 \\

\bottomrule 
\end{tabular}
\end{adjustbox}
\end{table}

\begin{figure}
\centering
  \includegraphics[width=0.35\columnwidth]{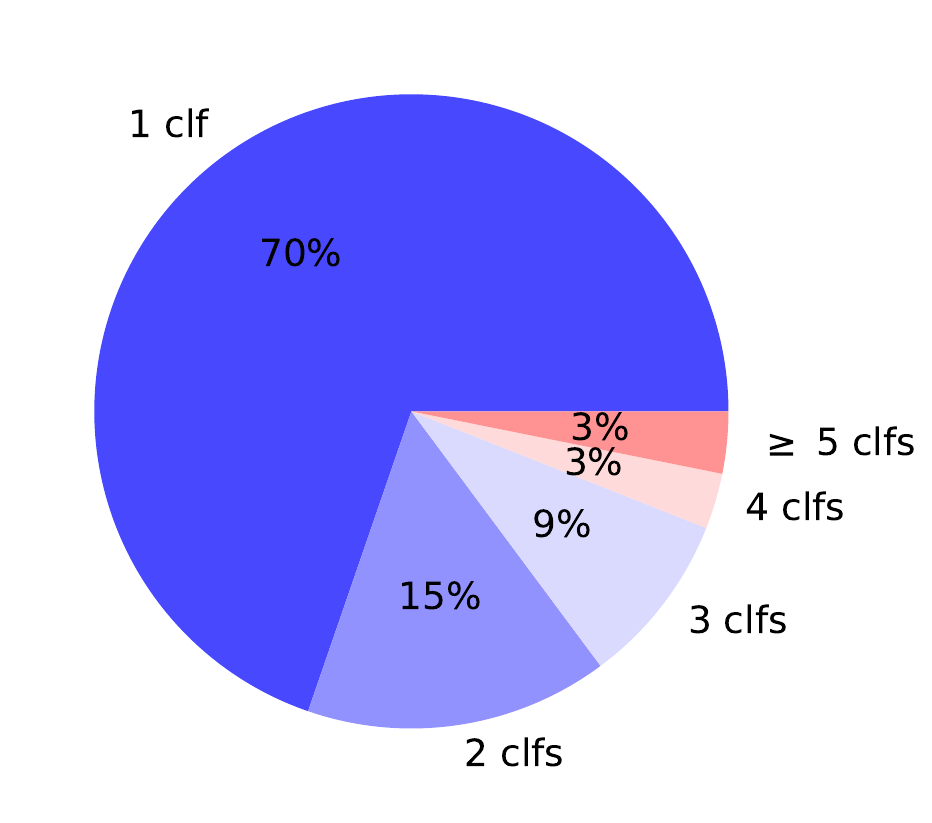}
  \caption{Number of classification models (clf) used for evaluation.}
  \label{fig:clfs}
\end{figure}

Figure \ref{fig:clfs} illustrates the number of different classification models considered during experiments. It is clear to see that the majority of publications (70\%) applied their bias mitigation method to only one classification model.
While in-processing methods are model specific and directly modify the training procedure,
pre-processing and most post-processing bias mitigation
methods can be developed independently from the classification models they are used for. Therefore, they can be devised
once and applied to multiple classification models for evaluating their performance.
Our observations confirm this intuition: only 24\% of publications with in-processing methods consider more than one classification model, while 35\% and 43\% of pre- and post-processing methods consider more than one respectively.

\section{Datasets}
\label{section:datasets}
In this section, we investigate the use of datasets for evaluating bias mitigation methods.
Among these datasets, some have been divided into multiple subsets (e.g., risk of recidivism or violent recidivism, medical data for different time periods).
For clarity, we treat data from the same source as a single dataset. 

Following this procedure, we gathered a total of 83 unique datasets.
We discuss these datasets in Section~\ref{data:freq} (e.g., what is the most frequently used dataset?)
and Section~\ref{data:count} (e.g., how many datasets do experiments consider?).
Additionally, 56 publications created synthetic or semi-synthetic datasets for their experiments.
Section~\ref{data:synth} provides information on the creation of such synthetic data.

For further details on datasets, we refer to \citet{le2022survey} who surveyed 15 datasets and provided detailed information on the features and dataset characteristics.
Additionally, \citet{kuhlman2020no} gathered 22 datasets from publications published in the ACM Fairness, Accountability, and Transparency (FAT) Conference and 2019 AAAI/ACM conference on Articial Intelligence, Ethics and Society (AIES).
Fairness datasets for a variety of domains (e.g., health, linguistics, social sciences, computer vision) can be found in the web app by \citet{fabris2022algorithmic}.\footnote{http://fairnessdata.dei.unipd.it/}

\subsection{Dataset Usage}
\label{data:freq}
In this section, we investigate the frequency with which each dataset set has been used.
The purpose of this analysis is to highlight the importance of each dataset and recommend the most important datasets to use for evaluating bias mitigation methods.
For this purpose, we consider \pubexp of the \puball publications, as only these \pubexp publications perform empirical experiments. The remaining publications do not present any empirical experiment and thus do not consider any dataset.

Among the 83 datasets, two are concerned with synthetic data (i.e., ``synthetic'' and ``semi-synthetic'') which we address in Section~\ref{data:synth}.
Therefore, we are left with \dataunique datasets.
59\% of the datasets (48 out of \dataunique) are used only once during experiments.
Another 14\% of the datasets (11 out of \dataunique) are only used twice. Thereby, 73\% of the datasets (59 out of \dataunique) are used rarely (by one or two publications). 

Table \ref{table:dataset-count} list the frequency of the remaining 22 datasets (used in three or more publications). A list of all datasets can be found in our online repository~\citep{homepage}.
In addition to the frequency, a percentage is provided (i.e., how many of the \pubexp publications use this datasets).
Among all datasets, the Adult dataset is used most frequently (by 77\% of the publications). While the Adult dataset contains information from the 1994 US census, \citet{ding2021retiring} derived new datasets from the US census from 2014 to 2018.

Five other datasets are used by 10\% or more of the publications (COMPAS, German Communities and Crime, Bank, Law School).
This shows that in order to enable a simple comparison with existing work, one should consider at least the Adult and COMPAS dataset.
However, these two datasets have recently received some criticism for their use as benchmark datasets and suitability as real-world datasets.
For instance, the Adult dataset applies a binary label to determine whether an individual has an income above 50,000 USD. \citet{ding2021retiring} showed that the fairness of ML models and bias mitigation methods is depending on the income threshold, thereby potentially limiting the external validity of the Adult dataset for benchmarking.
\citet{bao2021s} addressed the use of the Risk Assessment Instrument (RAI) datasets, in particular the COMPAS dataset, for benchmarking ML fairness.
They outlined that the use of such datasets should consider domain context, rather than using them as a generic example to show the real-world performance of bias mitigation methods.

\begin{table}[]
\caption{Frequency of widely used datasets (i.e., used in at least three publications).}
\centering
\label{table:dataset-count}
\begin{adjustbox}{max width=0.8\columnwidth}
\begin{tabular}{lrr}
\toprule
Dataset Name          & Frequency & Percentage \\
\midrule

 Adult~\citep{uci} & 249 & 77\% \\
 COMPAS~\citep{angwin2016machine} & 166 & 51\% \\
 German~\citep{uci} & 97 & 30\% \\
 Communities and Crime~\citep{redmond2002data} & 42 & 13\% \\
 Bank~\citep{moro2014data} & 38 & 12\% \\
 Law School~\citep{wightman1998lsac} & 33 & 10\% \\
 Default~\citep{yeh2009comparisons} & 24 & 7\% \\
 Dutch Census~\citep{dutchdata} & 16 & 5\% \\
 Health~\citep{healthdata} & 14 & 4\% \\
 MEPS~\citep{mepsdata} & 14 & 4\% \\
 Drug~\citep{fehrman2017five} & 9 & 3\% \\
 Student~\citep{cortez2008using} & 8 & 2\% \\
 Heart disease~\citep{uci} & 7 & 2\% \\
 National Longitudinal Survey of Youth~\citep{nlsy} & 6 & 2\% \\
 SQF~\citep{sqfdata} & 5 & 2\% \\
 Arrhythmia~\citep{uci} & 5 & 2\% \\
 Wine~\citep{cortez2009modeling} & 4 & 1\% \\
 Ricci~\citep{riccidata} & 4 & 1\% \\
 University Anonymous (UNIV) & 3 & 1\% \\
 Home credit~\citep{homecredit} & 3 & 1\% \\
 ACS~\citep{ding2021retiring} & 3 & 1\% \\
 MIMICIII~\citep{johnson2016mimic} & 3 & 1\% \\
\bottomrule 
\end{tabular}
\end{adjustbox}
\end{table}

\subsection{Dataset Frequency}
\label{data:count}
\begin{figure}
\centering
  \includegraphics[width=0.5\columnwidth]{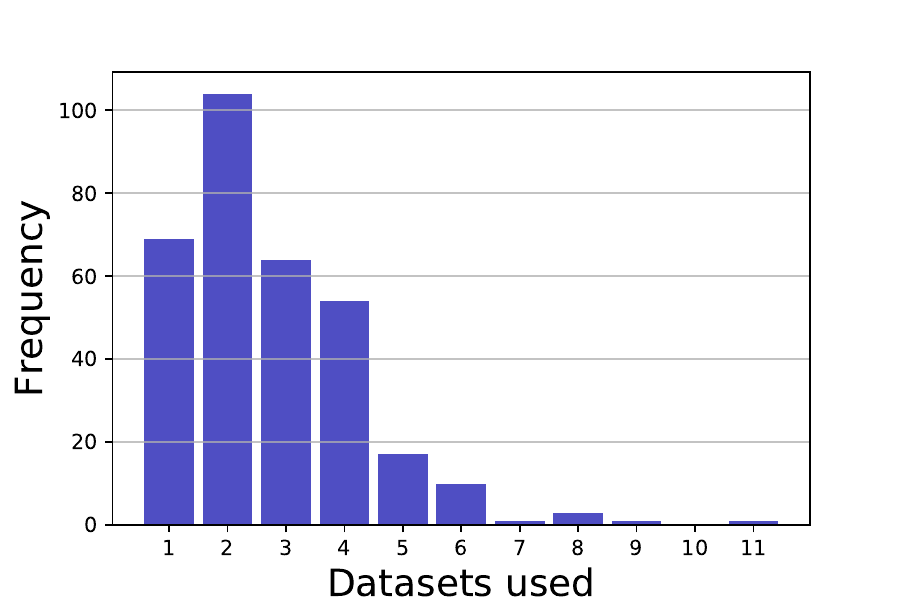}
  \caption{Number of datasets used per publication.}
  \label{fig:datasets}
\end{figure}

In addition to detecting the most popular datasets for evaluating bias mitigation methods, we investigate the number of different datasets used, as this impacts the diversity of the performance evaluation~\citep{kuhlman2020no}.
Figure \ref{fig:datasets} visualizes the number of datasets used for each of the \pubexp publications.

The most commonly used number of datasets considered for experiments is two, which
has been observed in 104 out \pubexp of the publications.
Overall, it can be seen that the number of considered datasets is relatively small (90\% of the publications use four or fewer datasets), with an average of \dataavg datasets per publication.
Two publications stand out in particular, with 9 datasets (\citet{fairsmote}), and 11 datasets (\citet{do2022fair}) respectively.
In accordance with existing work, new publications should evaluate their bias mitigation methods on three datasets, and if possible more.
Hereby it can be of interested to consider a diverse range of datasets based on application domains, dimensionality or protected attributes~\citep{le2022survey}.

\subsection{Synthetic Data}
\label{data:synth}
In addition to the \dataunique existing datasets for experiments, 54 publications created synthetic datasets to evaluate their bias mitigation method. 
Moreover, we found 3 publications that use semi-synthetic data (i.e., modify existing datasets to be applicable for evaluating bias mitigation methods) in their experiments~\citep{dwork2018decoupled,madras2019fairness,kilbertus2020fair}.

The created datasets range from hundreds of data points~\citep{dimitrakakis2019bayesian,olfat2018spectral,lahoti2019operationalizing,hashimoto2018fairness} to 100,000 and above~\citep{hickey2020fairness,iosifidis2019fairness,di2020counterfactual,zehlike2020matching}.
While the sampling procedures are well described, some publications do not state the dataset size used for experiments~\citep{zhang2018mitigating,mehrabi2022attributing,choi2021group,maity2020there,grari2021fairnessRenyi,balashankar2019fair,grabowicz2022marrying,kim2020fact}.

As exemplary data creation procedure, we briefly outline the data generation approach applied by \citet{zafar2017fairnesscons}, as it is the most frequently adapted approach by other publications~\citep{kilbertus2018blind,liu2022accuracy,zafar2017parity,roh2021sample,roh2020fr,roh2021fairbatch,Park2022privacy,kim2020fact}.
In particular, \citet{zafar2017fairnesscons} generated $4,000$ binary class labels. These are augmented with 2-dimensional user features which are drawn from different Gaussian distributions.
Lastly, the sensitive attribute is drawn from a Bernoulli distribution.

\subsection{Data-split}
\label{data:splits}
In this section we analyze whether existing publications provided information on the data-splits, in particular what sizing has been chosen.
Moreover, we investigate how often experiments have been repeated with such data splits, to account for training instability~\citep{friedler2019comparative}, therefore improving the conclusion validity of a study~\citep{qian2021my}.
Our focus lies on the data-splits used when evaluating the bias mitigation methods (e.g., we are not interested in data-splits that are applied prior for hyperparameter tuning of classification models~\citep{yan2020fair,oneto2020general,kehrenberg2019tuning,lahoti2019operationalizing,cotter2019two,li2022achieving,candelieri2022fair,huang2022fair,sikdar2022getfair}).

Among the \pubexp publications that carry out experiments, \setsplit provide information on the data-split used and \setuprun provide information on the number of \textit{runs} (different splits) performed.
The high amount of publications that do not provide information on the data-split sizes could be explained by the fact that some of the \dataunique datasets provided default splits.
For example, the Adult dataset has a pre-defined train-test split of 70\%-30\%, and \citet{cotter2019training} used designated data splits for four datasets.

A widely adopted approach for addressing data-splits for applying bias mitigation methods is k-fold cross validation. Such methods divide the data in $k$ partitions and use each part once for testing and the remaining $k-1$ partitions for training.
Overall, 47 publication applied cross validation: 10-fold (23 times), 5-fold (21 times), 3-fold (twice), 20-fold (once), and once without specification of $k$~\citep{goel2018non}.

If the data-splits are not derived from k-folds, the most popular sizes (i.e., train split size - test split size) are 80\%-20\% (39 times) and 70\%-30\% (35 times) followed by 67\%-33\% (16 times),  50\%-50\% (11 times), 60\%-40\% (5 times), and 75\%-25\% (5 times).
In addition to these regular sized datas-plits, there are 23 publication which divide the data into very ``specific'' splits. 
For example, \citet{quadrianto2018neural} divided the Adult dataset into $28,222$ training, $15,000$ and $2,000$ validation instance. 
Another example is the work by \citet{liu2022accuracy}, who chose $5.000$ training instances at random, using the remaining $40,222$ instances for testing.

Once the data is split in training and testing data, experiments are repeated 10 times in 54 out of \setuprun and 5 times in 42 out of \setuprun cases.
The most repetitions are performed in the work by \citet{da2020fairness}, who trained $48,000$ models per dataset to evaluate different hyperparameter settings.

We have found 16 publications that use different train and test splits for experiments on multiple datasets. 
Reasons for that can be found in the stability of bias mitigation methods when dealing with a large amount of training data~\citep{bechavod2017penalizing}.

While most publications split the data in two parts (i.e., training and test split), there are 36 publication that use validation splits as well.
The sizes for validation splits range from 5\% to 30\%, whereas the most common split uses 60\% training data, 20\% testing data, and 20\% validation data.
Furthermore, \citet{mishler2021fade} allow for a division of the data in up to five different splits for evaluating their ensemble learning procedure. 

Bias mitigation methods that process data in a streaming~\citep{zhang2019faht,iosifidis2020mathsf,iosifidis2019fairness,zhang2021farf,SHORT2022}, federated learning~\citep{ezzeldin2021fairfed,hu2022provably,qi2022fairvfl,pentyala2022privfairfl,abay2020mitigating}, multi-source~\citep{iofinova2021flea}, sequential~\citep{zhao2021fairness,zhao2022adaptive,Almuzaini2022abc,rateike2022don} fashion need to be addressed differently, as they use small subsets of the training data instead of using all at once.

\section{Fairness Metrics}
\label{section:metrics}

\begin{table*}[]
\caption{Popular fairness metrics. At least one metric for each category is provided.}
\centering
\label{table:metrics-frequent}
\begin{adjustbox}{max width=\columnwidth}
\begin{tabular}{lrrl}
\toprule
Name                          & Section & \# & Description                                                                    \\ \midrule
Statistical Parity Difference & \ref{section:def2}       & 136 & Difference of positive predictions per group                                   \\
Equality of Opportunity       & \ref{section:def3}         & 91 & Equal TPR per population groups                                                \\
Disparate Impact, P-rule      & \ref{section:def2}         & 60 & Ratio of positive predictions per group                                        \\ 
Equalized Odds                & \ref{section:def3}         & 51 & Equal TPR and FPR per population groups                                        \\
False Positive Rate           & \ref{section:def3}         & 38 & False positive rate difference per group                                       \\
Accuracy Rate Difference      & \ref{section:def3}         & 29 & Difference of prediction accuracy per group                                    \\
                              &    ...     &    & ...                                                                         \\
Causal Discrimination         & \ref{section:def5}         & 7  & Different predictions for identical individuals except for protected attribute \\
Mean Difference               & \ref{section:def1}         & 6  & Difference of positive labels per group in the datasets                        \\
Mutual information   & \ref{section:def6}         & 4  & Mutual information between protected attributes and predictions                \\
                              &    ...     &    & ...                                                                            \\
Strong Demographic Disparity  & \ref{section:def4}         & 1  & Demographic parity difference over various decision thresholds                \\
\bottomrule
\end{tabular}
\end{adjustbox}
\end{table*}

Fairness metrics play an integral part in the bias mitigation process.
First they are used to determine the degree of bias a classification model exhibits before applying bias mitigation methods. Afterwards, the effectiveness of bias mitigation methods can be determined by measuring the same metrics after the mitigation procedure.
In particular, this section focuses on metrics used for measuring bias, rather than general notions of fairness such as \textit{Fairness through Unawareness} (i.e., not using the protected attribute).


Recent fairness literature has introduced a variety of different fairness metrics, that each emphasize different aspect of classification performance.

To provide a structured overview of such a large amount of metrics, we devise metric categories, and take into account the classifications by \citet{caton2020fairness}, and \citet{verma2018fairness}.
Overall we categorize the metrics used in the \puball publications in six categories, which are defined based on  labels in dataset, predicted outcome,
predicted and actual outcomes, predicted probabilities and actual outcome, similarity, causal reasoning.

In the following, we provide information on how these metric types have been used.
In total, we found \metricsunique unique metrics that have been used by the \pubexp publications that performed experiments.
Most publications consider a binary setting (i.e., two populations groups and two class labels for prediction), whereas fairness has also been measured for non-binary sensitive attributes~\citep{alabdulmohsin2022reduction, zeng2022fair2,celis2021fairadv,celis2021fair,fairn2021sharma}, and multi-class predictions~\citep{alghamdi2022beyond,alabdulmohsin2022reduction}.

While some of the categories only contain few different metrics (definitions based on labels in dataset, on predicted probabilities and actual outcome, and on similarity all have 13 or fewer different metrics); \textit{definitions based on predicted outcome} have 22, \textit{definitions based on predicted and actual outcomes} have 31, and \textit{definitions based on Causal Reasoning} 27 different metrics.
Therefore, we outline the most frequently used metrics for \textit{definitions based on predicted and actual outcomes} and \textit{definitions based on causal reasoning}.

On average, publications consider two fairness metrics when evaluating bias mitigation methods, with 45\% of the publications only using one fairness metric. 
The most frequently used metrics are outlined in Table~\ref{table:metrics-frequent}, while listing at least one metric per category.
For detailed explanations of fairness metrics, we refer to \citet{verma2018fairness}.

In addition to quantifying the bias according to prediction tasks, we found metrics that determined fairness in accordance with feature usage (e.g., do users think this feature is fair~\citep{grgic2018beyond}) and quality of representations~\citep{samadi2018price,mcnamara2017provably,song2019learning} (see Section~\ref{section:pre-representation}).

\noindent
\textbf{Notations.} To provide equations of fairness metrics, we use the following notation:

\begin{compactitem}
    \item $S$: sensitive attribute to divide populations in two groups ($s_1$, $s_2$).
    \item $y$: Ground truth label.
    \item $\hat{y}$: Predicted label (or probability, Section~\ref{section:def4}).
    \item $Pr$: Probability.
    \item $D$: Dataset, with $N$ instances.
\end{compactitem}

\subsection{Definitions Based on Labels in Dataset}
\label{section:def1}
Fairness definition based on the dataset labels, also known as ``dataset metrics'', are used to determine the degree of bias in an underlying dataset~\citep{bellamy2018ai}.
One purpose of datasets metrics is determine whether there is a balanced representation of privileged and unprivileged groups in the dataset.
This is in particular useful for pre-processing bias mitigation methods, as they are able to impact the data distribution of the training dataset.

Most frequently, datasets metrics are used to measure the disparity in positive labels for population groups, such as Mean Difference (MD), elift and slift~\citep{morina2019auditing}, defined as follows:
\begin{align*} 
MD &= Pr(y = 1 | S = s_1) - Pr(y = 1 | S = s_2) \\
elift &= e^{-\epsilon} \leq  \frac{Pr(y = 1 | S = s)}{Pr(y = 1)}   \leq  e^{\epsilon} , \forall s \in S \\
slift &= e^{-\epsilon} \leq  \frac{Pr(y = 1 | S = s)}{Pr(y = 1 | S=s')}   \leq  e^{\epsilon} , \forall s, s' \in S
\end{align*}

\noindent elift and slift are parameterized by $\epsilon$, which allows for an easy comparison of bias between different classification models, by contrasting the magnitude of their $\epsilon$ values. Perfect fairness is achieved by $\epsilon = 0$. Among these,
MD is the most popular metric, used in six publications.


\subsection{Definitions Based on Predicted Outcome}
\label{section:def2}
Definitions based on predicted outcome, or ``Parity-based'' metrics, are used to determine whether different population groups receive the same degree of favour.
For this purpose, only the predicted outcome of the classification needs to be known.

The most popular approach for measuring fairness according to predicted outcome is the concept of \textit{Demographic Parity}, which states that privileged and unprivileged groups should receive an equal proportion of positive labels.
This can be done as by computing their difference (Statistical Parity Difference) or their ratio (Disparate Impact).
Similar to Disparate Impact, the p-rule compares two ratios of positive labels ($group_1 / group_2$, $group_2/group_1$) and 
Among those two ratios, the minimum value is chosen. 
The mathematical definition of these metrics is given below:

\begin{align*} 
\text{Statisitcal Parity Difference (SPD)} &= Pr(\hat{y} = 1 | S = s_1) - Pr(\hat{y} = 1 | S = s_2) \\
\text{Disparate Impact (DI)} &=  \frac{Pr(\hat{y} = 1 | S = s_1)}{Pr(\hat{y} = 1 | S = s_2)} \\
\text{P-rule} &= min\left (\frac{Pr(\hat{y} = 1 | S = s_1)}{Pr(\hat{y} = 1 | S = s_2)}, \frac{Pr(\hat{y} = 1 | S = s_2)}{Pr(\hat{y} = 1 | S = s_1)}  \right )
\end{align*}


If the direction of bias is of no interest (i.e., it is not important which group receives a favourable treatment), then the absolute bias values can be considered~\citep{PETROVIC2021104398,delobelle2020ethical,petrovic2022fair,raff2018gradient}.
While it is possible to compute fairness metrics based on differences as well as ratios between two groups, both which have been applied in the past, \citet{zliobaite2015survey} advised against ratios as they are more challenging to interpret.


\subsection{Definitions Based on Predicted and Actual Outcomes}
\label{section:def3}
Definitions based on predicted and actual outcomes are used to evaluate the prediction performance of privileged and unprivileged groups (e.g., is the classification model more likely to make errors when dealing with unprivileged groups?).
Similar to definitions based on predicted outcomes, the rates for privileged and unprivileged groups are compared.

Frequently, metrics based on predicted and actual outcomes are computed from combinations of confusion matrix measures (i.e., True Positives (TP), False Positives (FP), False Negatives (FN), True Negatives (TN)), as follows:

\begin{align*} 
\text{True Positve Rate (TPR)} &= \frac{TP}{TP+FN} \\
\text{False Positve Rate (FPR)} &= \frac{FP}{FP+TN} \\
\text{False Negative Rate (FNR)} &= \frac{FN}{FN+TP} \\
\text{True Negative Rate (TNR)} &= \frac{TN}{TN+FP} \\
\text{Positive Predictive Rate (PPR)} &= \frac{TP}{TP+FP} \\
\text{Negative Predictive Rate (NPR)} &= \frac{TN}{TN+FN} \\
\text{False Discovery Rate (FDR)} &= \frac{FP}{TP+FP} \\
\end{align*}

The most popular metric of this type is \textit{Equality of Opportunity} (used 90 times), followed by \textit{Equalized odds} (used 52 times). 
While \textit{Equality of Opportunity} is satisfied when populations groups have equal TPR, \textit{Equalized odds} is satisfied if population groups have equal TPR and FPR.
An average score of TPR and FPR is provided by the \textit{Average Odds Difference}. The formal definition of these metrics is shown below:

\begin{align*} 
\text{Equality of Opportunity} &= TPR_{S=s_1} -  TPR_{S=s_2} \\
\text{Equalized Odds} &= (FPR_{S=s_1} -  FPR_{S=s_2}) + (TPR_{S=s_1} -  TPR_{S=s_2}) \\
\text{Average Odds} &= \frac{1}{2} ((FPR_{S=s_1} -  FPR_{S=s_2}) + (TPR_{S=s_1} -  TPR_{S=s_2}))\\
\end{align*}

In addition to evaluating fairness in according to the confusion matrix (FPR - 38 times, TNR - 8 times), the accuracy rate (i.e., difference in accuracy for both groups) has been used 29 times.
Moreover, conditional TNR and TPR have been evaluated~\citep{salazar2021automated,salimi2019interventional} and one can compare populations groups with regards to performance metrics, such as precision, recall, F1 and Area Under Curve.

\subsection{Definitions Based on Predicted Probabilities and Actual Outcome}
\label{section:def4}
While Section \ref{section:def3} detailed metrics based on actual outcomes and predicted labels, this Section outlines metrics that consider predicted probabilities instead. 

\citet{jiang2020wasserstein} proposed Strong Demographic Disparity (SDD) and Strong Pairwise Demographic Parity  (SPDD), which are parity metrics computed over a variety of thresholds (i.e., prediction tasks apply a threshold of 0.5 by default):

\begin{align*} 
\text{Strong Pairwise Demographic Parity (SPDD)} &= \mathbb{E}_{\tau \sim U(\Omega))} | Pr(\hat{y} > \tau | S = s_1) - Pr(\hat{y} > \tau | S = s_2) | 
\end{align*}

\noindent where $\mathbb{E}_{\tau \sim U(\Omega))}$ denotes the expectation over all possible thresholds $\tau$, uniformly sampled from all possible prediction outcomes $U(\Omega)$.

\citet{chzhen2020fair} also varied thresholds, to compute the Kolmogorov-Smirnov distance. 
\citet{heidari2018fairness} measured fairness based on positive and negative residual differences.
\citet{agarwal2019fair} computed a Bounded Group Loss (BGL) to minimize the worst loss of any group, according to least squares.

Another notion of fairness based on predicted probabilities and actual outcomes is calibration~\citep{pleiss2017fairness}.
Calibration describes a scenario where predicted probabilities have a semantic meaning, for example if 100 individuals receive a prediction of $0.75$, then 75 of them should have a positive label (i.e., a label of $1$).
\citet{zhang2022longitudinal,zhang2021fair} proposed the use of a related metric with \textit{fair calibration} (FC).
FC first sorts predicted probabilities for each subgroup and divides them in 10 equally sized bins (e.g., 100 instances would result in 10 bins of 10 individuals).
It is then evaluated whether the 10 bins of each population group are calibrated, and in a second stage whether differences between predictions and actual outcomes are consistent across population groups.
FC then generates a binary result, whether the model is fairly calibrated or not.

\subsection{Definitions Based on Similarity}
\label{section:def5}
Definitions based on similarity are concerned with the fair treatment individuals. In particular, it is desired that individuals that exhibit a certain degree of similarity receive the same prediction outcome.
For this purpose, different similarity measures have been applied.
The most popular similarity metric used is \textit{consistency} or \textit{inconsistency} (used in 4 and 1 publications respectively)~\citep{zemel2013learning}.
\textit{Consistency} compares the prediction of an individual with the k-nearest-neighbors according to the input space~\citep{zemel2013learning}:

\begin{align*} 
\text{Consistency} &= 1 - \frac{1}{Nk}\sum_{n} |\hat{y}_n - \sum_{j\in kNN(x_n)} \hat{y}_j|
\end{align*}

\noindent \citet{luong2011k} also utilized k-nearest-neighbors, to investigate the difference in predictions for different values of $k$.

Similarities between individuals have been computed according to 
$\ell_\infty$-distance~\citep{ruoss2020learning}, and euclidean distance with weights for features~\citep{zemel2013learning}.
Individuals have also been treated as similar if they have equal labels~\citep{berk2017convex}, are equal except for sensitive features or based on predicted labels~\citep{verma2021removing}.
If similarity of individuals is determined solely by differences in sensitive features, one is speaking of ``causal discrimination''~\citep{lohia2019post,zhu2021learning}.\footnote{Some publications refer to this as ``Counterfactual fairness'~\citep{monteiroproposal,yurochkin2021sensei,Yurochkin2020Training}, but we follow the guidelines of \citet{verma2018fairness} and treat counterfactual fairness as a Causal metric.}

In contrast to determining similarity computationally, \citet{jung2019algorithmic} allowed stakeholders to judge whether two individuals should receive the same treatment.

Moreover, \citet{ranzato2021fairness} considered four types of similarity relations (\textsc{Noise}, \textsc{Cat}, \textsc{Noise-Cat}, \textsc{conditional-attribute}), when dealing with numerical and categorical features.
\citet{verma2021removing} considered two types of similarities: input space (identical on non-sensitive features), output space (identical prediction).
\citet{lahoti2019operationalizing} built a similarity graph to detect similar individuals. This graph is built based on pairwise information on individuals that should be treated equally with respect to a given task.


\subsection{Causal Reasoning}
\label{section:def6}
Fairness definitions based on causal reasoning take causal graphs in account to evaluate relationships between sensitive attributes and outcomes~\citep{verma2018fairness}.

For example, Counterfactual fairness states that a causal graph is fair, if the prediction does not depend on descendants of the protected attribute~\citep{kusner2017counterfactual}.
This definition has been adopted by four publications.
Moreover, the impact of protected attributes on the decision has been observed in two ways: direct and indirect prejudice~\citep{zhang2017causal}.
Direct discrimination occurs when the treatment is based on sensitive attributes. Indirect discrimination results in biased decision for population groups based on non-sensitive attributes, which might appear to be neutrals.
This could occur due to statistical dependencies between protected and non-protected attributes.

Direct and indirect discrimination can be modelled based on the causal effect along paths taken in causal graphs~\citep{zhang2017causal}. 
To measure indirect discrimination, Prejudice Index (PI) or Normalized Prejudice Index (NPI) haven been applied four times~\citep{kamishima2012fairness}.
NPI quantifies the mutual information between protected attributes and predictions. The mathematical definition of these measures follow:

\begin{align*} 
PI &= \sum_{y,s \in D}^{} Pr(y,s) ln \frac{Pr(y,s)}{Pr(s)Pr(y)} \\
NPI &= PI / (\sqrt{H(Y)H(S))}))  
\end{align*}

\noindent Here, $H(X)$ is defined as the entropy function $-\sum_{x \in D}^{} Pr(x) ln Pr(x)$.

Mutual information has also been used to determine the fairness of representations~\citep{moyer2018invariant,song2019learning}.
Similar to determining the degree of mutual information between sensitive attributes and labels, the ability to predict sensitive information based on representations has been used in nine publications.

\section{Benchmarking}
\label{section:benchmarking}
After establishing on which datasets bias mitigation methods are applied, and which metrics are used to measure their performance (Section \ref{section:metrics}), we investigate how they have been benchmarked.

Benchmarking is important for ensuring the performance of bias mitigation methods.
Nonetheless, we found 15 out of \pubexp publications that perform experiments but do not compare results with any type of benchmarking (i.e., out of the \puball publications, \pubexp perform experiments, among which \pubbench perform benchmarking).
Therefore, the remaining section addresses \pubbench publications which: 1) perform experiments; 2) apply benchmarking.

\begin{table}[]
\caption{Benchmarking against bias mitigation method types. For each bias mitigation category (i.e., pre-, in-, or post-processing), we count the type of benchmarking methods.}
\centering
\label{table:benchmark-other-bmm}
\begin{adjustbox}{max width=.4\columnwidth}
\begin{tabular}{lrrrrrr}
\toprule
                           &        &      & \multicolumn{3}{c}{Benchmarked against} \\ \cmidrule(l){3-6} 
                        Category    & \# & None & Pre    & In    & Post    \\ \midrule

 Pre  & 114  &  50 & 55  &  37  &   16  \\ 
 In   &   184  &  66 & 56  &  108  &   51      \\ 
 Post &   52  &  16 & 17  &  25  &   27        \\ 
\bottomrule
\end{tabular}
\end{adjustbox}
\end{table}

\subsection{Baseline}
\label{section:baseline}
To determine whether bias mitigation methods are able to reduce effectively, different types of baselines have been used. 
We use the term ``baseline'' to describe simple methods for benchmarking, that can be applied as a basic yet necessary check to determine whether a bias mitigation methods is effective. 
Unlike methods presented in Section~\ref{section:bmm} and Section~\ref{section:other_methods}, these are not based on existing methods from the \puball publications.

The most general baseline is to compare the fairness achieved by classification models after applying a bias mitigation method with the fairness of a fairness-agnostic \textit{Original Model}. 
If a method is not able to exhibit an improved fairness over a fairness-agnostic classification model, then it is not applicable for bias mitigation.
Given that this is the minimum requirement for bias mitigation methods, it is the most frequently used baseline (used in 254 out of \pubbench experiments).

Another baseline method is \textit{suppressing}, which performs a naive attempt of mitigating bias by removing the protected attribute from the training data.
However, it has been found that solely removing protected attributes does not remove unfairness~\citep{calders2009building,pedreshi2008discrimination}, as the remaining features are often correlated with the protected attribute.
To combat this risk, \citet{kamiran2010discrimination} suppressed not only the sensitive feature but also the k-most correlated ones.
\textit{Suppressing} has been used in 30 out of \pubbench experiments.

Random baselines constitute more competitive baselines than solely suppressing the protected attribute.
Bias mitigation methods that outperform random baselines show that they are not only able to improve fairness but also able to perform better than naive methods.
Random baselines have been used in 13 out of \pubbench experiments.

Moreover, we found four publications that considered a constant classifier for benchmarking (i.e., a classifier that returns the same label for every instance)~\citep{moyer2018invariant,mohammadi2022feta,wang2020robust,kim2020fact}. This serves as a fairness-aware baseline, as every individual and population group receive the same treatment~\citep{fairea}.

\subsection{Benchmarking Against Bias Mitigation Methods}
\label{section:bmm}
In addition to baselines, we investigate how methods are benchmarked against other, existing bias mitigation methods.
In particular, we are interested in which methods are popular, how many bias mitigation methods are used for benchmarking, and to what category these methods belong.

At first, we investigate what type of bias mitigation method are considered for benchmarking (e.g., are pre-processing methods more likely to benchmark against other pre-processing methods or in-/post-processing methods).
Table \ref{table:benchmark-other-bmm} illustrates the results. 
In particular, \# shows how many unique publications propose a given type of bias mitigation method (i.e., there are 114 publications with pre-processing methods).
For each of these methods we determine whether they benchmark against pre-, in- or post-processing methods. If no benchmarking against other bias mitigation methods is performed, we count this as ``None''.

We find that pre-processing methods are the most likely to not benchmark against other bias mitigation methods at 44\% (50 out of 114). 36\% (66 out of 184) of in-processing methods and 31\% (16 out of 52) of post-processing methods do not benchmark against other bias mitigation methods.
Furthermore, we can see that each bias mitigation type is more likely to benchmark against methods of the same type.

In addition to detecting the type of bias mitigation methods for benchmarking, we are interested in what approaches in particular are used for benchmarking.
Therefore, we count how often each of the \puball bias mitigation methods we gathered have been used for benchmarking.

\begin{figure}
\centering
\begin{tikzpicture} 

\begin{axis}[tickwidth = 0pt,xbar, xmin=0, width=0.5\columnwidth, height=12cm, enlarge y limits=0.2, 
ytick={1,2,3,4,5,6,7,8,9,10,11,12,13,14,15,16,17,18},
yticklabels={\citet{hardt2016equality},\citet{kamiran2012data},\citet{zhang2018mitigating},\citet{zafar2017fairnesscons},\citet{zafar2017fairness},\citet{agarwal2018reductions},\citet{feldman2015certifying},\citet{zemel2013learning},\citet{kamishima2012fairness},\citet{calmon2017optimized},\citet{madras2018learning},\citet{calders2010three},\citet{kamiran2010discrimination},\citet{donini2018empirical},\citet{pleiss2017fairness},\citet{kamiran2012decision},\citet{kamiran2009classifying},\citet{Louizos2016}}, 
ytick=data, nodes near coords,y axis line style = { opacity = 0 },
   axis x line       = none ] 
\addplot coordinates {(43,1)
(37,2)
(36,3)
(32,4)
(28,5)
(25,6)
(23,7)
(20,8)
(18,9)
(15,10)
(13,11)
(12,12)
(11,13)
(10,14)
(10,15)
(8,16)
(8,17)
(8,18)
}; 
\end{axis} 
\end{tikzpicture}
\caption{Most frequently benchmarked publications. For each publication, the number of times it has been used for benchmarking is shown.}
\label{fig:benchmarked}
\end{figure}
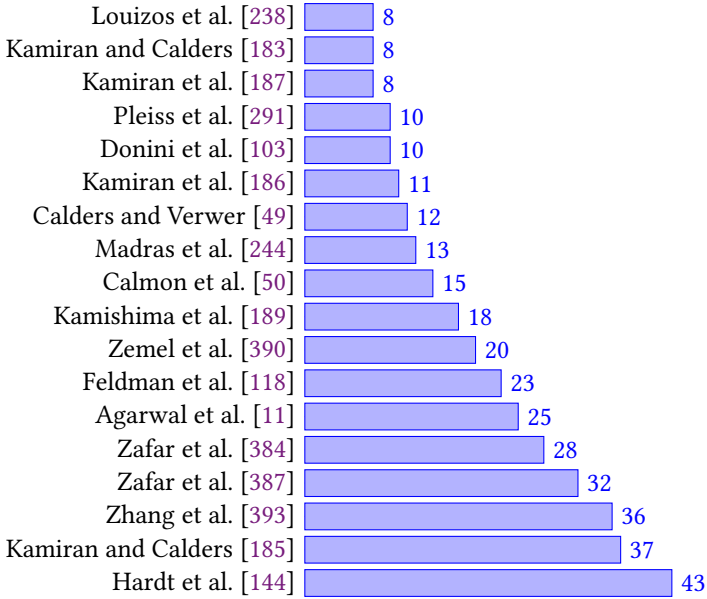

Overall, \benchunique bias mitigation methods have been used as a benchmark by at least one other publication.
Figure \ref{fig:benchmarked} illustrates the most frequently used bias mitigation methods for benchmarking.
Among the 18 listed methods, all of which are used for benchmarking by at least eight other publications, eight are pre-processing, nine in-processing, and four post-processing.
Notably, the five most-frequently used methods include each of the three types: sampling and relabelling for pre-processing~\citep{kamiran2012data}, constraints~\citep{zafar2017fairnesscons,zafar2017fairness} and adversarial learning~\citep{zhang2018mitigating} for in-processing, and classifier modification for post-processing~\citep{hardt2016equality}.

\subsection{Benchmarking Against Fairness-Unaware Methods}
\label{section:other_methods}
In addition to benchmarking against existing bias mitigation methods, practitioners can use other methods for benchmarking, which are not designed for taking fairness into consideration.
Overall, we found \benchother publications that use fairness-unaware methods for benchmarking (i.e., using a general data augmentation method to benchmarking fairness-aware resampling).

Table \ref{table:benchmark-other-methods} shows the publications that benchmark their proposed method against at least one fairness-unaware methods, according to the type of approach applied.
Among the 13 types of approaches, as shown in Section~\ref{section:pre} - \ref{section:post}, seven can be found to benchmark against fairness-unaware methods.
This occurs rarely for post-processing methods, six publications in total, with at least one per approach type.
A total of 23 and 27 publications for pre-processing and in-processing methods, respectively, benchmark against fairness-unaware methods.

\begin{table}[]
\caption{Publications that benchmark against at least one fairness-unaware method.}
\centering
\label{table:benchmark-other-methods}
\begin{adjustbox}{max width=0.9\columnwidth}
\begin{tabular}{llrp{32em}}
\toprule
Type                  & Category       & Section & References                                                                                                                                     \\ \midrule
\multirow{2}{*}{Pre}  & Sampling       & \ref{section:pre-sampling}       & \makecell[l]{
\citet{cruz2021bandit, celis2020data, xu2019fairgan, abusitta2019generative} \\
\citet{du2021robust, yan2020fair, roh2021sample, xu2019achievingan} \\
\citet{zhang2021farf, dablain2022towards, pentyala2022privfairfl}
} \\ 
                      & \cellcolor{gray!25} Representation & \cellcolor{gray!25} \ref{section:pre-representation}        & \cellcolor{gray!25} \makecell[l]{\citet{Louizos2016, creager2019flexibly, salazar2021automated, gupta2021controllable} \\ \citet{balunovic2022fair, Galhotra2022causal, qi2022fairvfl, oh2022learning} \\
                      \citet{lahoti2019ifair, jaiswal2020invariant,shui2022fair, sarhan2020fairness}}   \\ \midrule
\multirow{4}{*}{In}   & Regularization & \ref{section:in-loss}        & \citet{liu2021fair, zhang2021fair, jiang2022generalized, zhang2022longitudinal,wang2022synthesizing}                                                                                   \\ 
                      & \cellcolor{gray!25} Constraints    & \cellcolor{gray!25} \ref{section:in-loss}        & \cellcolor{gray!25} \makecell[l]{
\citet{zhao2021fairness, ding2020differentially, du2021robust, zhang2021omnifair} \\
\citet{fukuchi2015prediction, narasimhan2018learning, zhao2022adaptive,wang2022synthesizing} 
}                   \\  
                      & Adversarial    & \ref{section:in-adversarial}        & \makecell[l]{
\citet{xu2019fairgan, lahoti2020fairness, roh2020fr, sadeghi2019global} \\
\citet{yazdani2022distraction,rezaei2021robust} 
}                                                \\ 
                      &  \cellcolor{gray!25} Adjusted     & \cellcolor{gray!25} \ref{section:in-adjusted}        &
                      \cellcolor{gray!25}  \makecell[l]{\citet{cruz2021bandit, luo2015discrimination, iosifidis2020mathsf, liu2021fair} \\ \citet{fairn2021sharma, zhang2021farf, candelieri2022fair, wang2022mitigating} \\
                      \citet{zhao2022adaptive, maheshwari2022fairgrad,lee2021fair}}
                                             \\ \midrule
\multirow{3}{*}{Post} & Input          & \ref{section:post-in}       & \citet{adler2018auditing}                                                                                                                       \\ 
                      &   \cellcolor{gray!25} Classifier  & \cellcolor{gray!25} \ref{section:post-clf}       &\cellcolor{gray!25}  \citet{mehrabi2022attributing, wu2022fairness}                                                                                                                 \\
                      &   Output        & \ref{section:post-out}       & \citet{kamiran2018exploiting, alabdulmohsin2021near, pentyala2022privfairfl}              \\ \bottomrule                                                                                                   
\end{tabular}
\end{adjustbox}
\end{table}


\subsection{Source Code Availability}
To investigate whether existing work allows for reproducibility of the results and ease of use for benchmarking, we reviewed whether the \puball surveyed publications shared source code.
Specifically, we have collected links to implementations from the publications directly. If no link was available, we performed a google search to check for resources we might have missed.\footnote{For each publications, we searched for ``\textit{paper title}'' and ``\textit{paper title github}'' and checked the first page of search results for links to external resources.}
With this additional search, we were able to find 64 implementations.
Overall, we found 192 publications with available source code (56\% of the \puball publications).

\begin{figure}
\centering
  \includegraphics[width=0.7\columnwidth]{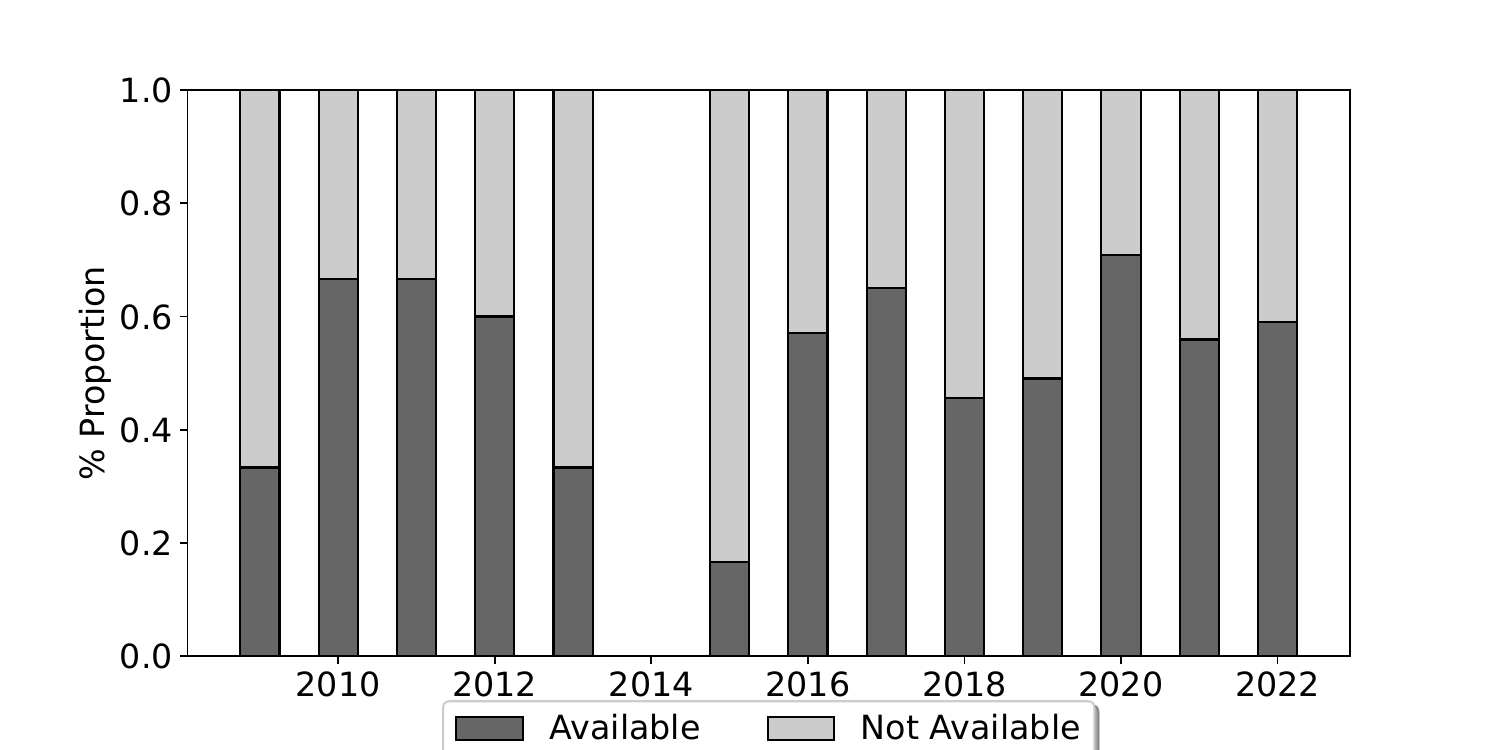}
  \caption{Proportion of publications that publicly shared the source code used in their study, per year.}
  \label{fig:codeAvail}
\end{figure}

Figure~\ref{fig:codeAvail} illustrates the proportion of publications with code available per year.
Early years (2009-2016) show a high variation in the proportion of publications with source code available, ranging from 17\% to 67\%.
Such a variation is caused by the small number of publications.
In 2018 and 2019, the proportion of publications with shared source code is below 50\%, 46\% and 49\% respectively. The most recent years showed an increase in shared implementations, with the maximum achieved in 2020 with 71\% of the publications to share source code.

Moreover, we examined existing surveys for frameworks providinf implementations of bias mitigation methods~\citep{chen2022fairness, soremekun2022software,pessach2022review,dunkelau2019fairness,mehrabi2021survey,caton2020fairness,lee2021landscape}, and found three frameworks that do so: Themis-ML~\citep{bantilan2018themis}, AIF 360~\citep{bellamy2018ai}, Fairlearn~\citep{bird2020fairlearn}.
In total, Themis-ML~\citep{bantilan2018themis} implements bias mitigation from three publications, Fairlearn~\citep{bird2020fairlearn} implements four methods, and AIF 360~\citep{bellamy2018ai} implements 13 methods.\footnote{1st of March 2023}

While our focus lies on the sharing and reuse of bias mitigation methods, datasets are also an important resource to share to allow for reproducibility.
Many datasets are already publicly available, however some datasets are proprietary and cannot be shared publicly. Where available, we provide links to datasets and source code implementations in our  online repository~\citep{homepage}.

\section{Challenges and Opportunities}
\label{section:discussion}
This section provides further discussion and insights on the surveyed publications. We outline several challenges, based on the current literature, as well as discuss research opportunities for the creation and evaluation of new bias mitigation methods.

\subsection{Challenges}\label{section:challengens}
Research on bias mitigation is fairly young and does therefore enable challenges and opportunities for future research.
Here, we highlight five challenges that we extracted from the collected publications, that call for future action or extension of current work.

\subsubsection{Fairness Definitions}
A variety of different metrics have been proposed and used in practice (see Section~\ref{section:metrics}), which can be applied to different use cases.
However, with such a variety of metrics it is difficult to evaluate bias mitigation on all and ensure their applicability.
Consolidating a common set of metrics to use is still an open challenge~\citep{garcia2020reducing,mehrabi2021survey,dablain2022towards}, as can be seen by the use of \metricsunique different fairness metrics in the literature, as discussed in Section~\ref{section:metrics}.
While consolidating existing fairness notions is one problem, it is also relevant to ensure that the used metrics are representative for the problem at hand\cite{baresiREnext,SarroRE23}.
Often, this means evaluating fairness in a binary classification problem for two population groups. While this can be the correct way to model fairness scenarios, it is not sufficient to handle all cases, such that future work should focus on multi-class problems~\citep{kamiran2012data,monteiroproposal,morina2019auditing,celis2019improved,huang2022fair}, and non-binary sensitive attributes, which was mentioned by only 15 publications~\citep{kamiran2012data,kamishima2012fairness,menon2018cost,grari2021fairness,ravichandran2020fairxgboost,valdivia2021fair,kang2021multifair,petrovic2022fair,baharlouei2019r,celis2019improved,feng2019learning,calders2010three,kamiran2009classifying,feldman2015certifying,alabdulmohsin2022reduction}.

Other challenges regarding metrics include the trade-offs when dealing with accuracy and/or multiple fairness metrics~\citep{agarwal2018reductions,pessach2020algorithmic,caton2020fairness,obermeyer2019dissecting}, as well as the allowance of some degree of discrimination as long it as explainable (e.g., enforcing a fairness criteria completely could lead to unfairness in another)~\citep{calders2010three,kamiran2012data,kamiran2010classification,zemel2013learning}.
    


\subsubsection{Fairness Guarantees}
Guarantees are of particular importance when dealing with domains that fall under legislation and regulatory controls~\citep{feldman2015certifying,kamishima2012fairness}.
Thereby, it is not always sufficient to establish the effectiveness of a bias mitigation method based on the performance on the test set without any guarantees.
Fairness guarantees can help in this situation, by providing performance guarantees with regards to a specific fairness metric and bound the degree of bias~\citep{celis2019classification,joseph2016fairness}.
In particular, \citet{dunkelau2019fairness} pointed out that most bias mitigation methods are evaluated on test sets and their applicability to real-world tasks depends on whether the test set reliably represents reality.
If that is not the case, fairness guarantees could ensure that bias mitigation methods are able to perform well with respect to a given fairness metric and unknown data distributions. 
Therefore, eight publications considered fairness guarantees as a relevant avenue of future work. 
Similarly, allowing for interpretable and explainable methods can aid in this regard~\citep{woodworth2017learning,johndrow2019algorithm,quadrianto2018neural,kamishima2012fairness}.


\subsubsection{Datasets}
Another challenge that arises when applying bias mitigation methods is the availability and use of datasets.
The most pressing concern is the reliability and access to protected attributes, which was mentioned in nine publications, as this information is often not available in practice~\citep{holstein2019improving}.

Moreover, it is not guaranteed that the annotation process of the training data is bias free~\citep{hardt2016equality}.
If possible an unbiased data collection should be enforced~\citep{quadrianto2017recycling}.
Other options are the debiasing of ground truth labels~\citep{yu2021fair,zhu2021learning} or use of expert opinions to annotate data~\citep{du2021fairness}.
If feasible, more data can be collected~\citep{johndrow2019algorithm,chen2018my}, which is difficult from a research perspective, as commonly, existing and public datasets are used without the chance to manually collect new samples.

Besides, the variety of protected attributes addressed in previous experiments, as found by \citet{kuhlman2020no}, is lacking diversity, with the majority of cases considering race and gender only.
In practice, ``collecting more training data'' is the most common approach for debiasing, according to interviews conducted by \citet{holstein2019improving}.
However, an interviewee questioned whether such a fairness intervention is fair, as the targeting of subgroups for additional data collection may be a biased procedure.

    
\subsubsection{Real-world Applications}
While the experiments are conducted on existing, public datasets, it is not clear whether they can be transferred to real-world applications without any adjustments.
For example, \citet{hacker2017continuous} see the challenge of data distributions changing over time, which would require continuous implementations of bias mitigation methods.

Moreover, developers might struggle to detect the relevant population groups to consider when measuring and mitigating bias~\citep{holstein2019improving}, whereas the datasets investigated in Section~\ref{section:datasets} often simplify the problem and already provide binarized protected attributes (e.g., in the COMPAS, six ``demographic'' categories are transformed to ``Caucasian'' and ``not Caucasian''~\citep{bellamy2018ai}). 
Therefore, \citet{martinez2020minimax} stated that automatically identifying sub-populations with high-risk during the learning procedure as a field of future work.

Given the multitude of fairness metrics (as seen in Section~\ref{section:metrics}), real world applications could even suffer further unfairness after applying bias mitigation methods due to choosing incorrect criteria~\citep{lee2022maximal}.
Similarly, showing low bias scores does not necessarily lead to a fair application, as the choice of metrics could be used for ``Fairwashing'' (i.e., using fake explanations to justify unfair decisions)~\citep{mehrabi2022attributing,anders2020fairwashing}.
Nonetheless, \citet{sylvester2020trimming} argue that considering fairness criteria while developing ML models is better than considering none, even if the metric is not optimal.

\citet{sharma2020data} show the potential of user studies to not only provide bias mitigation methods that work well in a theoretical setting, but to make sure practitioners are willing to use them. In particular, the are interesting in finding how comfortable developers and policy makers are with regards to training data augmentation.

To facilitate the use and implementation of existing bias mitigation methods, metrics and datasets, popular toolkits such as AIF360~\citep{bellamy2018ai} and Fairlearn~\citep{bird2020fairlearn} can be used.

\subsubsection{Extension of Experiments}
Lastly, a challenge and field of future research is the extension of conducted experiments to allow for more meaningful results.

The most frequently discussed aspect of extending experiments is the consideration of further metrics (in 40 publications).
Moreover, the usefulness of bias mitigation methods can be investigated when applied to additional classification models. This was pointed out by 12 publications. 
Given the \dataunique datasets that were used at least once, and on average \dataavg datasets used per publication, only eight publications see the consideration of further datasets as a useful consideration for extending their experiments~\citep{krasanakis2018adaptive,chakraborty2020fairway,hort2021did,da2020fairness,yan2020fair,chakraborty2019software,candelieri2022fair,wang2021fair}.

While the consideration of additional metrics, classification models and datasets does not lead to changes in the training procedure and experimental design, there are also intentions to apply bias mitigation methods to other tasks and contexts, such as recommendations~\citep{kang2021multifair,zafar2017fairnesscons}, ranking~\citep{huang2019stable,kamishima2012fairness,zafar2017fairnesscons} and clustering~\citep{kamishima2012fairness}.

    


\subsection{Research Opportunities}
In the course of this survey, we have collected \puball publications with regards to various approaches for bias mitigation methods. 
This collection helps us understand which approaches have already been applied and allows us to outline some aspects that appear underexplored and provide opportunities for future research.

Firstly, from the \puball publications we collected, it can be seen that in-processing methods are the most widely explored methods. 
There are almost twice as many publications with in-processing methods than pre-processing, and nearly four times as many in-processing methods than post-processing methods.
Therefore, addressing post-processing bias mitigation method seems unexplored in contrast to the other two method types.
In particular the modification of inputs in a post-processing stage has only been considered by two publications (Section~\ref{section:post-in})~\citep{adler2018auditing,li2022training}.
However, this type of bias mitigation method could be further investigated without considerable effort by developing new methods, simply by applying existing pre-processing methods (Section~\ref{section:pre}) to the testing data.

Generally speaking, pre- and post-processing methods are classifier-agnostic and can be evaluated on a variety of classification models without modification to the underlying algorithm.
Nonetheless, Bandits have been investigated with neither of these two method types, only by in-processing methods~\citep{joseph2016fairness,joseph2018meritocratic,gillen2018online}.

Moreover, the combination of pre- and post-processing methods has only been addressed four times~\citep{zhang2018achieving,wei2020optimized,pentyala2022privfairfl,li2022training}.
The number of classification models considered by these four publications range from 1 to 3.
This is a promising combination of approaches, as one can perform experiments with bias mitigation methods at two different stages (i.e., before and after training) on various classification models and thereby collect extensive empirical evidence for fairness improvements.
Additionally, we found several publications that applied multiple bias mitigation methods of the same type (e.g., two pre-processing methods). Six of these applied multiple pre-processing methods and 19 applied multiple in-processing methods (Table~\ref{table:combined}).
However, we found no publication that applied multiple post-processing methods.

Lastly, our data collection shows that there exist a multitude of datasets and metrics, which can enable a rigorous evaluation of novel bias mitigation methods.

For one, bias mitigation methods can be evaluated on up to \dataunique datasets, whereas bias mitigation methods evaluated on three datasets exceed the average of \dataavg datasets used for evaluation.  When applying bias mitigation methods to a dataset, it is important to mention the protected attributes considered and potential criticisms that could impact the ability to make claims about applicability for real world systems~\citep{ding2021retiring,bao2021s}.

The \metricsunique metrics used in the literature thus far, are classified in six categories. Thereby, bias mitigation method can be evaluated by multiple metrics of a same category, or multiple metrics from different categories. In addition to using fairness metrics to evaluate the performance of bias mitigation methods, performance metrics, such as accuracy, should be used to determine the fairness-accuracy trade-off achieved when applying bias mitigation methods \cite{fairea}. To ensure the competitiveness of results, methods must always be benchmarked against baselines as well as previous existing relevant methods, especially when their implementation is made publicly available (our survey highlights that 192 studies provided source code implementations, and as such they could be used as a benchmark for future proposals).

\section{Current Best Practices / Recommendations}
\label{section:recommendations}
In this section, we would like to outline current practices for the empirical evaluation of bias mitigation methods, that we have observed from the \puball publications.
However, we note that increasing the comprehensiveness of the empirical evaluation is always positive to support the validity of results (e.g., applying bias mitigation methods to a higher number of datasets, or using more metrics for evaluation).
Our recommendations, which will allow new experiments to be in line with prior experiments conducted, are as follows:
\noindent
\textbf{1.} Check existing approaches, to confirm the novelty of the bias mitigation method under evaluation.

\noindent
\textbf{2.} Apply your bias mitigation method to at least three datasets, taking diversity and criticism into account when making claims about real world impact.

\noindent
\textbf{3.} State the protected attributes for each dataset.

\noindent
\textbf{4.} Evaluate your bias mitigation method on at least two fairness metrics, as well as an performance metric (e.g., accuracy). We suggest using different metric types to reduce the correlation of individual fairness metrics.

\noindent
\textbf{5.} Benchmark at least against the original model and consider similar, existing bias mitigation methods as well.

\noindent
\textbf{6.} Apply your bias mitigation method to multiple classification models, in particular when proposing pre- or post-processing methods. Logistic regression and neural networks are frequently used.

\noindent
\textbf{7.} Try to repeat experiments at least 10 times for standard training splits (e.g. 70\% or pre-defined data-splits).

\noindent
\textbf{8.} Share code and numerical results, in particular when results are presented in bar charts.

\section{Conclusion}\label{section_conclusion}
In this literature survey, we focused on the adoption of bias mitigation methods to achieve fairness in classification problems and provided an overview of \puball publications.
Our survey first categories bias mitigation methods according to their type (i.e., pre-processing, in-processing, post-processing). We found \pubpre pre-processing, \pubin in-processing, and \pubpost post-processing methods, showing that in-processing methods are the most commonly used.
We devised 13 categories for the three method types, based on their approach (e.g., pre-processing methods can perform sampling).
The most frequently applied approaches perform changes to the loss function in an in-processing stage (51 publications applying regularization and 74 applying constraints). 
Other approaches are less frequently used, with input correction in a post-processing stage only being used twice. 

We further provided insights on the evaluation of bias mitigation methods according to three aspects: datasets, metrics, and benchmarking.
We found a total of \dataunique datasets that have been used at least once by one of the \puball publications, among which the Adult dataset is the most popular (used by 77\% of publications). Even though \dataunique datasets are available for evaluating bias mitigation methods, only \dataavg datasets are considered on average.

Similarly, we found a large number of fairness metrics that have been used at least once (\metricsunique unique metrics), which we divide in six categories. The most frequently used metrics belong to two categories: 1) Definitions based on predicted outcome; 2) Definitions based on predicted and actual outcomes.

When it comes to benchmarking bias mitigation methods, they can be compared against baselines, other bias mitigation methods, or non-bias mitigation approaches. 
Among the three baselines we found (original model, suppressing, random), the 82\% of bias mitigation methods consider the original model (i.e., the classification model without any bias mitigation applied) as a baseline.
Commonly, methods are compared against other bias mitigation methods.
\benchother publications benchmark against fairness-unaware methods.
Among the collected publications, we found 56\% (192 out of \puball) that make their source code available, thereby supporting replicability and benchmarking. Moreover, we found three frameworks implementing and making available existing bias mitigation methods~\citep{bantilan2018themis,bellamy2018ai,bird2020fairlearn}.

Lastly, we list current opportunities and challenges that have been discerned from the collected publications.
This includes the synthesizing of fairness metrics, as there is no consensus reached on what metrics to use.
In addition to measuring improvements, future bias mitigation methods can take fairness guarantees in account.
The application of bias mitigation methods in practice is challenging, as developers might not be able to detect relevant population groups for which to measure bias and reliability of datasets (i.e., are prior observations biased?).
Therefore, we hope that this survey helps researchers and practitioners to gain an understanding of the current, existing bias mitigation approaches and support the development of new methods.

\section*{Acknowledgements}
Zhenpeng Chen, Mark Harman and Federica Sarro are supported by the ERC grant no. 741278 (EPIC). Jie M. Zhang is partially supported by the UKRI Trustworthy Autonomous Systems Node in Verifiability, with Grant Award Reference EP/V026801/2. Max Hort is supported through the ERCIM ‘Alain Bensoussan’ Fellowship Programme.
We would like to thank the members of the community who kindly provided comments and feedback on an earlier version of this article.

\bibliographystyle{ACM-Reference-Format}
\bibliography{references}
\end{document}